\pdfoutput=1
\documentclass{article}
\usepackage{iclr2027_conference,times}
\usepackage{amsmath,amssymb}
\usepackage{booktabs}
\usepackage{graphicx}
\usepackage{xcolor}
\usepackage[utf8]{inputenc}
\usepackage{microtype}
\usepackage{url}
\usepackage{hyperref}
\hypersetup{colorlinks=true,linkcolor=black,citecolor=black,urlcolor=black,pdftitle={Same Answer, New Box: How Much of RLVR's Out-of-Domain Gain Is Answer Format?},pdfauthor={Zongye Lyu}}

\iclrfinalcopy
\AddToHook{env/table/begin}{\setlength{\belowcaptionskip}{4pt}}

\title{Same Answer, New Box: How Much of RLVR's\\Out-of-Domain Gain Is Answer Format?}

\author{Zongye Lyu \\
Faculty of Information Technology, Monash University, Melbourne, VIC 3800, Australia \\
\texttt{lyuzongye@gmail.com}}

\newcommand{\grep}{g_{\mathrm{rep}}}
\newcommand{\gfix}{g_{\mathrm{fix}}}

\begin{document}
\maketitle
\lhead{Preprint}

\begin{abstract}
Reinforcement learning with verifiable rewards (RLVR) is credited with gains beyond the domain its rewards check,
such as higher MMLU accuracy after training on mathematics. Such accuracies are usually read out of the model's text
by an answer extractor. A model that learns to write its answer where the extractor looks can therefore score
higher without answering better. We measure how much of the gain reported by extractors copied from public code
arises this way. We generate outputs from a base model and its RL model and score them twice: with those
extractors, and with one extractor of our own that reads a stated answer whether it is in the answer box
\verb|\boxed{}| or in the text. On four released 7B SimpleRL base--RL pairs, 82\% of the out-of-domain
multiple-choice gain that these extractors report disappears under ours: it is a gain in answer format, not in
answers. The base models already state most of their answers in text, and RL moves them into the box, where these
extractors look first. Independent LLM readers, in place of our extractor, put the figure at 77\%. With a prompt
that itself asks for the box, our extractor puts it at 79\%. On a 32B pair trained with another recipe, almost none of the
reported gain is format gain: its base model boxes its answers when asked, and its RL model rarely does when not. A reported gain is partly a
property of the extractor that read it, and re-scoring stored outputs with a second extractor shows how much,
without re-running RL.
\end{abstract}

\section{Introduction}

Reinforcement learning with verifiable rewards (RLVR) trains a language model on problems whose answers a program
can check, mostly mathematics and code. Its released models are also compared with their base models on benchmarks
outside that domain, such as MMLU, and a gain there is read as a sign that what the model learned transfers
\citep{simplerlzoo,barriers,mathtransfer}.

Such accuracies are usually read out of the generated text. A program, the \emph{answer extractor}, finds the answer
in the text and compares it with the key, and evaluation and training code each ship their own. We call the
extractors copied from such public code \emph{harness extractors}, and the gain that one of them reports for an RL
model over its base model, published or not, a \emph{reported gain}. If RL changes where a model writes its answer, a
harness extractor can report a gain that the answers do not contain.

\begin{figure}[t]
\centering
\includegraphics[width=\textwidth]{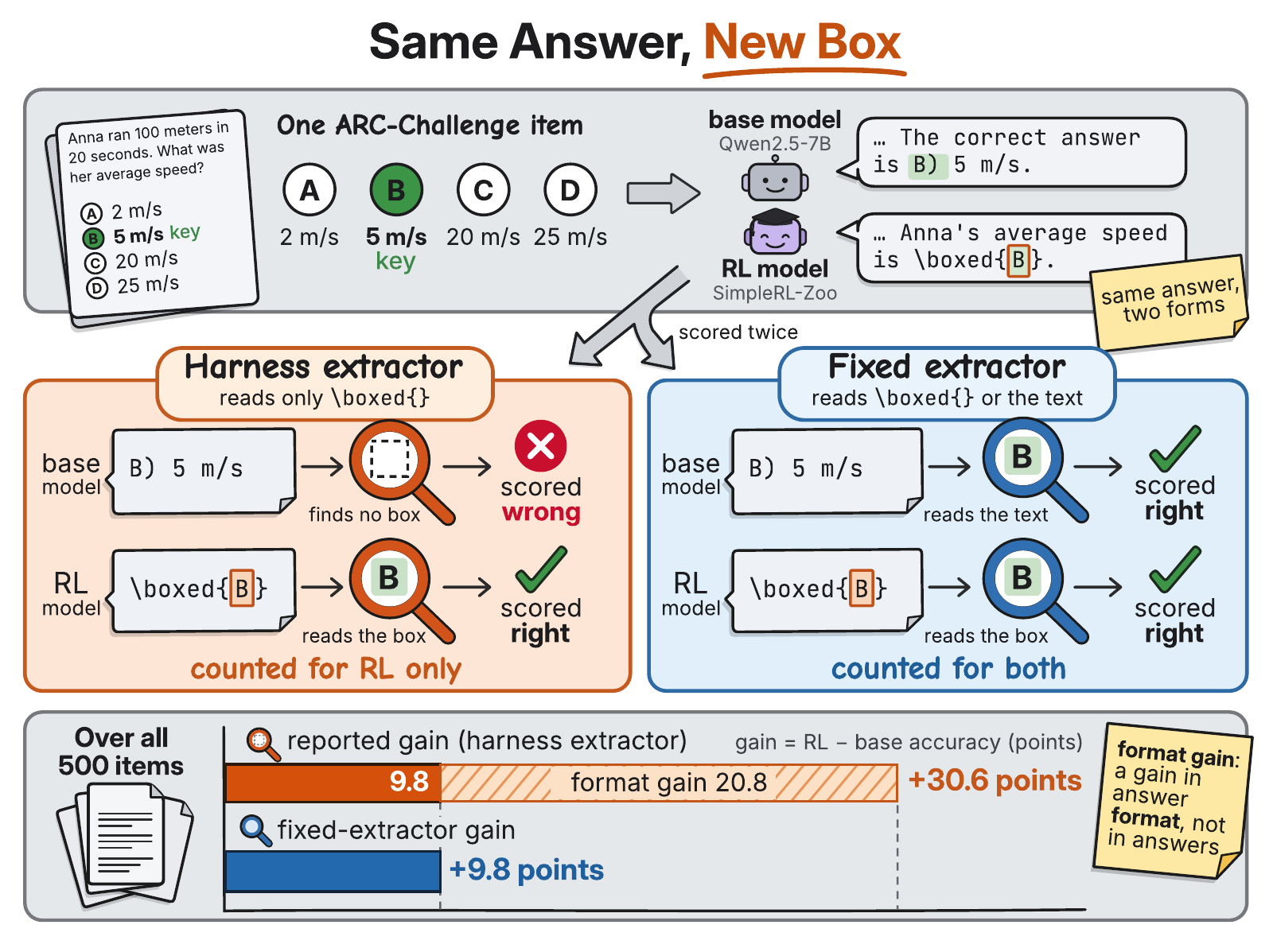}
\caption{\textbf{The same answer, counted for one model only.} One ARC-Challenge item and the end of each model's
output: Qwen2.5-7B (base) and the model trained from it by RL in the SimpleRL-Zoo release (RL) both answer B. A
harness extractor that reads only \texttt{\textbackslash boxed\{\}} (left) does not read the base model's answer
and scores it wrong; our fixed extractor (right) reads a stated answer in either form. Bottom: over the 500
ARC-Challenge items we evaluate, the harness extractor reports that RL raised accuracy by 30.6 points, and the fixed
extractor reads 9.8 from the same outputs. The difference, 20.8 points, is the format gain.}
\label{fig:example}
\end{figure}

Figure~\ref{fig:example} shows this on one question. Asked an ARC-Challenge question, the base model Qwen2.5-7B
ends with ``The correct answer is B) 5 m/s.'' Its RL-trained successor from the SimpleRL-Zoo release
\citep{simplerlzoo} ends with ``Anna's average speed is \verb|\boxed{B}|.'' The two answers are the same, and both
are correct. An extractor that reads only \verb|\boxed{}| counts the second and misses the first. Over the 500
ARC-Challenge questions we evaluate, that extractor reports that RL raised accuracy by 30.6 points. An extractor that
finds a stated answer in either form reads a gain of 9.8 points from the same outputs.

A gain on an out-of-domain benchmark can therefore mean two different things. If the RL model gives more correct
answers, what it learned transfers. If it gives the same answers but writes more of them where the extractor looks,
the benchmark has rewarded a change of format, not better answers.

That RL can learn a format is known; how much of a reported gain the format accounts for, on released models, has
not been measured. On MATH-500, a reward for merely producing a \verb|\boxed{}| expression recovers about half of the
gain of the true reward \citep{spurious}, and in RLVR with one training example a reward for format alone recovers
most of it \citep{oneshot}. On their own RL runs, \citet{fixedsae} find that RL mostly adds formatting scaffolding and
that what transfers from GSM8K to MATH-500 is mostly format. An essay by \citet{baselines} compares the base-model
accuracies that seven RL papers report with official ones and argues that part of the reported gains comes from
evaluations that under-read the base model, answer format and parsing among the causes. None of these takes a
released checkpoint's own outputs, with no weights changed, and measures how much of the gain reported on them is
produced by the extractor. To our knowledge, this paper is the first to do so.

The two kinds of gain can be told apart on outputs that already exist. We re-read the same outputs of a base model
and its RL model with one \emph{fixed extractor}, the same for both models, which reads a stated answer whether it is
in \verb|\boxed{}| or in the text. The part of the reported gain that this extractor does not find is the
\emph{format gain}, and its fraction of the reported gain is the \emph{format share} (Section~\ref{sec:method}). We
apply this to released base--RL pairs, with harness extractors copied from six public repositories
(Section~\ref{sec:setup}), and find:

\textbf{Most of the out-of-domain multiple-choice gain is format gain.} On MMLU, MMLU-Pro and ARC-Challenge, 82\% of
the gain that harness extractors report for four released 7B SimpleRL base--RL pairs is format gain
(Figure~\ref{fig:gain}).

\textbf{RL moves the stated answer into \texttt{\textbackslash boxed\{\}}.} The base models state most of their
answers in the text, and the RL models write most of theirs in \verb|\boxed{}|. In in-domain free-form mathematics, where
both already box their answers, an extractor that reads only \verb|\boxed{}| shows a format gain close to zero
(Figure~\ref{fig:position}).

\textbf{Most of the gain remains format gain under our other choices and under independent readers.} The share moves
by a few points when the fixed extractor reads more or less strictly, when one pair or benchmark is left out, or when
only one of our two prompts is used. The 82\% pools only the comparisons with a clearly positive reported gain
(Section~\ref{sec:criteria}); pooling every comparison whose reported gain is not zero lowers it the most, to 75\%.
Independent LLM readers of the whole responses, in place of the fixed extractor, find 77\% (Figure~\ref{fig:robust}).

\textbf{A reported gain depends on its extractor.} On identical outputs, the largest and smallest gains that harness
extractors report for the same pair, benchmark and prompt differ by a median of 15 points. An extractor that also
reads letters in the text reports much less format gain (Figure~\ref{fig:extractors}).

These are the gains that harness extractors report on outputs we generated from the released checkpoints. The two
prompts are ours, and so is the use of the extractors that read \verb|\boxed{}| first on these benchmarks, which
their repositories do not evaluate. They are not the numbers a release publishes, whose own evaluations we did not reproduce. They are also
gains of 7B models: on the one larger pair with gains to split, a 32B model trained with another recipe, almost none
of the reported gain is format gain (Section~\ref{sec:limits}). What does carry over to published numbers is how they
are produced: where the releases we surveyed say how, their out-of-domain multiple-choice results were almost always
read out of generated text (Section~\ref{sec:disagree}), so a format gain can arise in them too.

Our contribution is this measurement: how much of the out-of-domain multiple-choice gain that harness extractors
report for released RLVR checkpoints is format gain. It needs only the two models' outputs, re-read by one extractor
on a CPU, and no RL run.

\section{Splitting a reported gain}
\label{sec:method}

We read the same outputs twice: once with a harness extractor, whose gain is the reported one, and once with a fixed
extractor that treats the two models alike. What the harness extractor reports beyond the fixed extractor's gain is
the format gain.

\subsection{Two gains from the same outputs}

In Figure~\ref{fig:example}, the 20.8 points between the harness extractor's 30.6 and the fixed extractor's 9.8 are
gain that exists only for the extractor that looks in the box. We call them the \emph{format gain},
and their fraction of the reported gain, here 68\%, the \emph{format share}.

In general, fix a base--RL pair, a benchmark and a prompt, and let $h$ be the harness extractor and $\omega$ the fixed
one. Each extractor gives each model an accuracy, and the two gains are
\begin{equation}
\grep = \mathrm{acc}(\text{RL}, h) - \mathrm{acc}(\text{base}, h), \qquad
\gfix = \mathrm{acc}(\text{RL}, \omega) - \mathrm{acc}(\text{base}, \omega).
\end{equation}
The format gain is $D = \grep - \gfix$, and the format share is $D/\grep$. Because $\omega$ applies the same rules
to both models' outputs, $\gfix$ is the gain in the answers as $\omega$ reads them, and $D$ is the rest.

We call one pair on one benchmark under one prompt, scored by one harness extractor, a \emph{comparison};
Figure~\ref{fig:example} shows one. Over many comparisons we sum before dividing, so the pooled format share is
$\sum D / \sum \grep$. This weights each comparison by its reported gain, so a comparison with a gain near zero
cannot dominate the result.

\subsection{The fixed extractor}
\label{sec:omega}

The fixed extractor has one job: to count the same answer the same way for both models. It
therefore reads an answer only where one is explicitly stated: inside \verb|\boxed{}|, after ``the answer is'' or
``Answer:'', on a line of its own, or after ``option'' or ``choice''. Appendix~\ref{app:omega} gives its rules in
full, and two looser versions of it.

Where it finds no answer, it abstains, and the item counts as wrong for either model. It never drops an item,
because an extractor allowed to skip items could raise its own accuracy by skipping the hard ones. It abstains
about equally often on the two models: on the comparisons behind the 82\%, it reads an answer in 87\% of the base
models' outputs and 88\% of the RL models'. Section~\ref{sec:robust} checks what its abstentions cost each model.

\subsection{Which comparisons we pool}
\label{sec:criteria}

A format share is a fraction of a gain, so it means something only where there is a gain to split and where the
fixed extractor can read both models. We therefore pool the comparisons that meet two criteria. The harness
extractor must report a clearly positive gain, so that the ratio never divides by noise. And the fixed extractor must
find an answer in at least half of each model's outputs, at rates within 20 points of each other, so that its gain is
not read from a few outputs of one model. (The share of a model's outputs in which an extractor finds an answer is that
extractor's \emph{extraction rate} on the model.) Table~\ref{tab:grid} gives the thresholds, and Section~\ref{sec:robust}
changes both. Our 95\% intervals resample items within each benchmark. They do not cover variation between pairs or
between benchmarks, which we show instead by leaving each one out.

\section{Setup}
\label{sec:setup}

To see what RL alone changes, we compare released pairs in which the RL model was trained from a named public base
model by RL with verifiable rewards, with no supervised fine-tuning or distillation in between. We study seven such
pairs from three releases (Table~\ref{tab:grid}): SimpleRL-Zoo \citep{simplerlzoo}, SimpleRL-Zero from the same group,
and Open-Reasoner-Zero \citep{orz}. Their base models are Qwen2.5-Math-7B, Qwen2.5-7B and Qwen2.5-1.5B \citep{qwen25math,qwen25},
Mistral-7B-v0.1 \citep{mistral} and DeepSeek-Math-7B \citep{deepseekmath}. Twelve other candidates fail this rule. Three
more pairs meet it, two larger pairs and a Llama-3.1-8B pilot, and Section~\ref{sec:limits} reports them
(Appendix~\ref{app:population}).

\begin{table}[t]
\caption{\textbf{What we ran.} Every model answers every item under both prompts, and every harness extractor that
reads the benchmark's kind of answer scores the same outputs. The 82\% pools the out-of-domain multiple-choice
comparisons of the four pairs marked $\ast$ that meet both criteria in the last two rows; the other three pairs show
no clear out-of-domain multiple-choice gain that the fixed extractor can read (Section~\ref{sec:setup}).}
\label{tab:grid}
\centering
\small
\begin{tabular}{p{0.2\textwidth}p{0.74\textwidth}}
\toprule
base--RL pairs & Qwen2.5-Math-7B with SimpleRL-Zoo$^{\ast}$ and with SimpleRL-Zero$^{\ast}$; Qwen2.5-7B with
SimpleRL-Zoo$^{\ast}$; Mistral-7B-v0.1 with SimpleRL-Zoo$^{\ast}$; DeepSeek-Math-7B with SimpleRL-Zoo; Qwen2.5-1.5B
with SimpleRL-Zoo and with Open-Reasoner-Zero \\ \addlinespace
benchmarks & out of domain: MMLU, MMLU-Pro and ARC-Challenge (multiple choice), BBH (free-form); in domain: MATH-500
and GSM8K (free-form), AQuA-RAT and SAT-Math (multiple choice); up to 500 items each \\ \addlinespace
harness extractors for multiple choice, and where each looks & MATH evaluation code: the last
\texttt{\textbackslash boxed\{\}}\newline
Open-Reasoner-Zero: the last \texttt{\textbackslash boxed\{\}}\newline
Qwen2.5-Math: the last \texttt{\textbackslash boxed\{\}} first; if that gives no letter, a letter from the text
(on MMLU, the letter first)\newline
simple-evals: a line ``Answer: X''\newline
(five more read free-form answers only; Appendix~\ref{app:extractors}) \\ \addlinespace
prompts & boxed-instruction: asks for the final answer inside \texttt{\textbackslash boxed\{\}}, on multiple choice
as a single letter; plain: no format request of its own. Greedy decoding, at most 2,048 new tokens \\
\midrule
significant-gain criterion & the reported gain's lower 95\% bound is at least 2 points \\ \addlinespace
extraction-rate criterion & the fixed extractor reads an answer in at least 50\% of each model's outputs, and the two
rates are within 20 points (a comparison that meets it is \emph{readable}) \\
\bottomrule
\end{tabular}
\end{table}

The claim concerns three multiple-choice benchmarks outside the domain that the rewards check: MMLU \citep{mmlu},
MMLU-Pro \citep{mmlupro} and ARC-Challenge \citep{arc}. To tell whether a format gain goes with the domain or with
the answer form, we add benchmarks for the other three combinations of the two. Out of domain, BBH \citep{bbh}
supplies free-form tasks; in domain, MATH-500 \citep{math,math500} and GSM8K \citep{gsm8k} supply free-form
mathematics, and AQuA-RAT \citep{aqua} and SAT-Math \citep{agieval} multiple choice. The items were frozen before any output was
generated.

The harness extractors come from six public repositories: the MATH evaluation code \citep{math}, Open-Reasoner-Zero,
Qwen2.5-Math \citep{qwen25math}, simple-evals \citep{simpleevals}, lm-evaluation-harness \citep{lmeval} and verl
\citep{verl}. The four that read multiple choice look for the answer in different places (Table~\ref{tab:grid}): two
read only the last \verb|\boxed{}|, Qwen2.5-Math's reads \verb|\boxed{}| first and then letters in the text (on
MMLU, the letter first), and simple-evals' looks for a line ``Answer: X''. We run Qwen2.5-Math's extractor as its repository does, with the \emph{letter step} of its evaluation loop, which reads a
letter from the text when the extraction function returns none. We keep a verbatim copy of each extractor at a pinned commit and score
with an executable version tested against it (Appendix~\ref{app:extractors}). We apply every multiple-choice
extractor to all three benchmarks. Where an extractor reads \verb|\boxed{}| first, this pairing is ours: its
repository does not use it on these benchmarks.

Each model answers each item under two prompts that differ in one sentence. On multiple-choice benchmarks the
\emph{boxed-instruction prompt} adds ``Reason step by step, then give your final answer as a single letter inside
\verb|\boxed{}|''; the \emph{plain prompt} adds no request for a format. We use both to see whether a format gain
remains when the prompt itself asks for the box. Both are sent in the checkpoint's chat template, where it ships one,
and keep the template's default system message. For Qwen2.5-Math-7B and its two RL models that message itself asks
for the final answer ``within \verb|\boxed{}|'', so for these two pairs the plain prompt also carries one request for
the box.

Four of the seven pairs show a clear out-of-domain multiple-choice gain that the fixed extractor can read, and the
82\% pools them: Qwen2.5-Math-7B with its SimpleRL-Zoo and SimpleRL-Zero successors, and Qwen2.5-7B and
Mistral-7B-v0.1, each with SimpleRL-Zoo. On our outputs the DeepSeek-Math-7B pair shows no such gain to split, and
the fixed extractor cannot read the Qwen2.5-1.5B base model, whose outputs almost all run to the token limit
(Appendix~\ref{app:population}).

\section{Results}
\label{sec:results}

\subsection{Most of the out-of-domain multiple-choice gain is format gain}
\label{sec:main}

\begin{figure}[t]
\centering
\includegraphics[width=\textwidth]{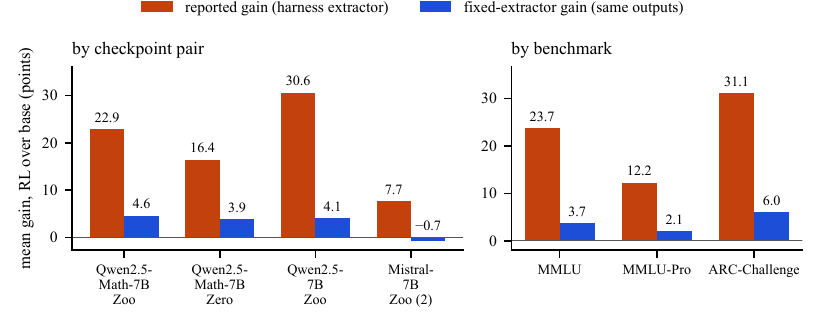}
\caption{\textbf{Most of the reported gain disappears when the same outputs are read by one fixed extractor.} Mean
gain of the RL model over its base model per comparison, on out-of-domain multiple choice, as the harness extractors
report it (orange) and as the fixed extractor reads the same outputs (blue), by base--RL pair and by benchmark. Each
bar averages several comparisons; that the Qwen2.5-7B Zoo bar reports 30.6, as Figure~\ref{fig:example} does, is a
coincidence. Zoo and Zero are the SimpleRL-Zoo and SimpleRL-Zero successors; Mistral-7B contributes 2 comparisons.
Pooled over the 46 comparisons, the format share is 82.2\% (95\% CI 77.4--87.2\%); it is between 81\% and 84\% on
each benchmark and between 76\% and 87\% for each Qwen pair.}
\label{fig:gain}
\end{figure}

We first ask how much of a reported gain is left when one extractor reads both models. Little is left
(Figure~\ref{fig:gain}). On MMLU, MMLU-Pro and ARC-Challenge, the harness extractors report that RL raised accuracy
by 22.5 points per comparison on average; read by the fixed extractor, the same outputs show a gain of 4.0 points.
Pooled, 82\% of the reported gain is format gain, and it is at least 76\% on each benchmark and for each of the three
Qwen pairs. Most of what these extractors report as an out-of-domain gain is a change in where the answers are
written, not better answers.

This holds for benchmarks that are both out of domain and multiple choice. On the study's other benchmarks the share
is not stable enough to quote: it moves by 50 points or more with the rule for which comparisons are pooled
(Appendix~\ref{app:strata}). The study therefore cannot say whether the format gain goes with
out-of-domain benchmarks or with multiple-choice answers.

\subsection{RL moves the stated answer into {\normalfont\ttfamily\textbackslash boxed\{\}}}
\label{sec:position}

\begin{figure}[t]
\centering
\includegraphics[width=\textwidth]{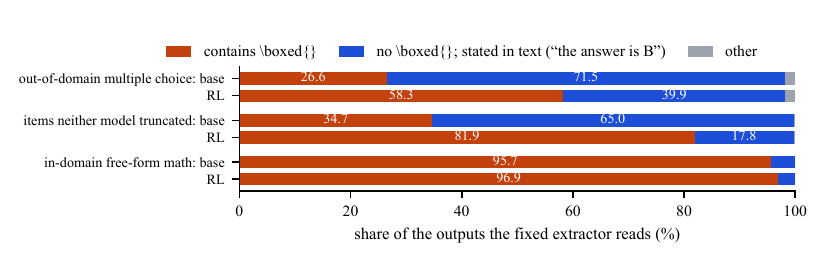}
\caption{\textbf{The base models state most of their answers in the text; the RL models box them.} Each bar divides
the outputs in which the fixed extractor reads an answer: those that contain a \texttt{\textbackslash boxed\{\}},
those with none whose answer is stated in the text, and other (a line holding only the letter, or an answer after
``option'' or ``choice''). Out of domain, an extractor that reads only \texttt{\textbackslash boxed\{\}} cannot see
most of the base models' answers; in in-domain free-form mathematics both models box almost every answer. Both prompts are pooled, over every pair
whose two models receive the same input, which is all but Open-Reasoner-Zero's, whose RL model ships a different chat
template. Out of domain, the four pairs of the 82\% alone show the same pattern, slightly stronger
(Appendix~\ref{app:population}).}
\label{fig:position}
\end{figure}

Why does a harness extractor see a gain that the fixed extractor does not? Because the two models write their
answers in different places (Figure~\ref{fig:position}). Among the out-of-domain multiple-choice outputs in which the
fixed extractor reads an answer, 26.6\% of the base models' contain a \verb|\boxed{}|, against 58.3\% of the RL
models'. Nearly all of the base models' other answers are stated in the text, after ``the answer is'' or
``Answer:'', with no \verb|\boxed{}| anywhere in the output. An extractor that reads only \verb|\boxed{}| therefore
misses most of the base models' answers and fewer of the RL models', and that difference is the format gain.

The token limit could also leave answers unboxed, but it does not explain the difference. An output cut off before
its end has no final \verb|\boxed{}|. Yet on the items where neither model reached the limit, the difference between
the two models is wider, not narrower: 34.7\% against 81.9\% (Figure~\ref{fig:position}, middle rows).

In-domain free-form mathematics is the contrast. On MATH-500 and GSM8K, 96\% to 97\% of the outputs in which the fixed
extractor reads an answer contain a \verb|\boxed{}|, for both models. The MATH and Open-Reasoner-Zero extractors,
which read the last \verb|\boxed{}|, then see what the fixed extractor sees, and their format gain is close to zero.

\subsection{Most of the gain remains format gain under our other choices and under independent readers}
\label{sec:robust}

\begin{figure}[t]
\centering
\includegraphics[width=\textwidth]{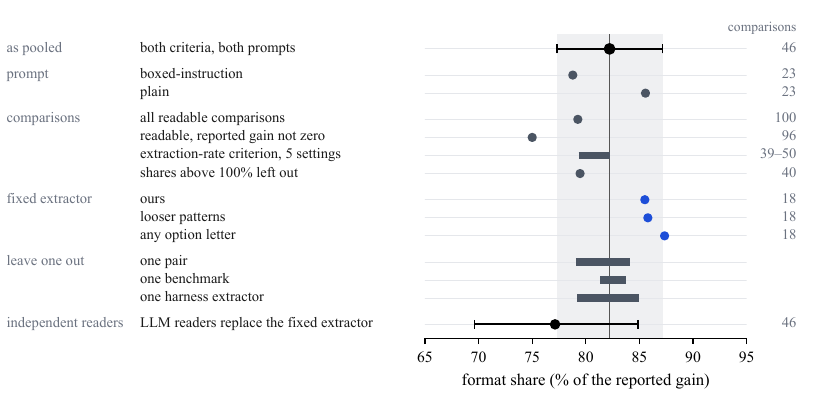}
\caption{\textbf{The format share under each of our choices, and under independent readers.} Each row changes one
thing from the pooled result at the top (the band is its 95\% interval) and recomputes the share on the comparisons
that remain, counted on the right. A share above 100\% means that the fixed extractor reads a loss. The rows on the
fixed extractor use the 18 comparisons that all three of its versions can read, on which our version gives 85.5\%. Table~\ref{tab:robust} in Appendix~\ref{app:robust} gives every value and the conditions of each
row.}
\label{fig:robust}
\end{figure}

The 82\% rests on choices we made and on an extractor we wrote. We therefore changed the choices shown in
Figure~\ref{fig:robust}, one at a time, and then replaced the extractor with independent readers. In every row of
Figure~\ref{fig:robust} most of the reported gain remains format gain: the share stays between 75\% and 87\%.

Most choices move the share by a few points. How strictly the fixed extractor reads, how the extraction-rate
criterion is set, and which pair or benchmark is left out each move it by at most 3.2 points
(Appendix~\ref{app:robust}). Which comparisons are pooled moves the share most: pooling every comparison with a non-zero
reported gain lowers it to 75\% (Appendix~\ref{app:robust}).

The prompt matters too, and in the direction the mechanism predicts. A base model that is asked for \verb|\boxed{}|
uses it more often, so fewer of its answers are hidden from an extractor that reads only the box, and less of the gain
is format gain. The share is 79\% under the boxed-instruction prompt and 86\% under the plain one. Even when the
prompt asks for the box, the RL models box more of their answers than the base models do, and most of the gain is
still format gain. A third prompt, from a pilot, gives a much lower share on one pair and benchmark
(Section~\ref{sec:limits}).

\label{sec:readers}The fixed extractor itself could be producing the result, so we also had the responses read by
readers that share no code with it. Two LLM readers each read 2,720 responses sampled from the outputs behind the
82\%, the same items for a base model and its RL model, and a third settled their disagreements
(Appendix~\ref{app:readers}). Each saw the question, its options and one response, and never the model, the key or
any extractor's reading. Scored by their readings, 77\% of the reported gain is format gain.

The readers and the fixed extractor read the same answer almost everywhere the extractor reads one: in 98.6\% of
those responses. They differ where it abstains. In about a tenth of the responses, for base and RL models alike, the
readers find an answer where the fixed extractor finds none. Those answers are more often correct for the RL models,
so the readers credit RL with a larger gain than the fixed extractor does, and a smaller part of the reported gain is
left as format gain.

By either reading, then, most of the reported gain is format gain. Our sample of responses is too small to say
whether 77\% and 82\% differ (Appendix~\ref{app:readers}).

\subsection{A reported gain depends on its extractor}
\label{sec:disagree}

\begin{figure}[t]
\centering
\includegraphics[width=\textwidth]{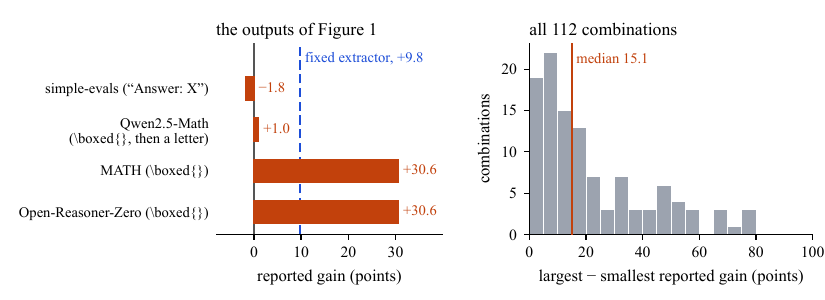}
\caption{\textbf{Different extractors report different gains on the same outputs.} Left: the gain each
multiple-choice harness extractor reports on the outputs of Figure~\ref{fig:example} (Qwen2.5-7B and SimpleRL-Zoo,
ARC-Challenge, boxed-instruction prompt, 500 items); the dashed line is the fixed extractor's reading. Right: over
all 112 combinations of pair, benchmark and prompt, the largest minus the smallest gain that the harness extractors
report; the median is 15.1 points. On each extractor's own comparisons in the 82\%, the format share is 85\% for
the MATH extractor, 83\% for Open-Reasoner-Zero's and 56\% for Qwen2.5-Math's (Table~\ref{tab:app-extractor-shares}).}
\label{fig:extractors}
\end{figure}

If a format gain comes from where the answers are written, the gain should change with the extractor that reads
them. It does, and sometimes even in sign (Figure~\ref{fig:extractors}). On the outputs of Figure~\ref{fig:example},
the two extractors that read only \verb|\boxed{}| report a gain of 30.6 points, and Qwen2.5-Math's, which also
takes letters from the text, reports 1.0. It reads base-model answers that the fixed extractor misses there, mostly an
option written out on a line of its own (``B) The dog becomes ill, and the tapeworm benefits.''), so on these outputs
the fixed extractor's 9.8 overstates the gain in answers (Appendix~\ref{app:omega}).

Simple-evals' extractor, which looks for the ``Answer:'' line that its own prompt asks
for and ours do not, reports a loss of 1.8: the RL models write their answers in \verb|\boxed{}|, and in most of the
readable out-of-domain comparisons that it scores it gives them exactly 0\%. Across all pairs, benchmarks and
prompts, the largest and smallest gains reported on the same outputs differ by a median of 15 points, and by 10
without simple-evals' two extractors.

An extractor that also reads letters in the text therefore reports less format gain. Of the gain that
Qwen2.5-Math's extractor reports on its own comparisons, 56\% is format gain; for the two extractors that read only
the box it is 85\% and 83\% (Appendix~\ref{app:robust}). Across the seven combinations of base model and harness
extractor that contribute comparisons to the 82\%, the share is lower where the extractor's extraction rate on the
base model is higher. This rests mostly on Qwen2.5-Math's extractor, which reads the most of the base models' outputs
and reports the least format gain (Appendix~\ref{app:tracks}).

How are published out-of-domain multiple-choice numbers produced? The 18 recent RLVR and reasoning-model releases
that we surveyed report 17 such results: 13 were computed by reading the answer out of generated text, one by
scoring log-likelihoods, and three do not say how (Appendix~\ref{app:survey}). A format gain can arise wherever the
answer is read out of text.

\section{Related work}
\label{sec:related}

\paragraph{RL's contribution and its transfer.} \citet{yue} find that RLVR-trained models beat their base models in
pass@$k$ at small $k$ while the base models score higher at large $k$, and conclude that current RLVR does not elicit
fundamentally new reasoning patterns. \citet{barriers} find that the benefits of RL post-training are largely
limited to the domains represented in its data, and \citet{mathtransfer} that RL-tuned models transfer across domains
better than supervised ones. These studies ask what RL adds and whether it transfers; we ask how much of a measured
gain the evaluation's extractor produces.

\paragraph{Measurement of RLVR gains.} \citet{sober} show that reasoning benchmarks are highly sensitive to
implementation choices and warn that formatting artifacts can produce spurious gains. \citet{position} argue that
headline RLVR gains conflate policy improvement with measurement confounds, and \citet{evalallyouneed} find that small
differences in evaluation conditions move results substantially. \citet{baselines} examine seven RL papers, find
the base-model accuracies they report below official ones in most, name answer format and parsing among the causes,
and ask for sample-level outputs to be released. On the input side, \citet{reachability} and \citet{passk}
control the prompt format given to each model and find that part of a gain remains when they do. \citet{ttra} first
adapt every model, the base model included, to the benchmark's task format and distribution with test-time RL, which
changes the weights. The gap between fine-tuned models, SimpleRL-Zoo among them, and their base models then largely
disappears. We hold the outputs fixed and change only the extractor.

\paragraph{Format in RL training.} \citet{spurious} show that RL with a reward for merely producing a
\verb|\boxed{}| expression raises Qwen2.5-Math-7B's MATH-500 accuracy by 13.8 points, about half of what the
ground-truth reward gives. \citet{fixedsae} track RL checkpoints with one sparse autoencoder and find that RL mainly
strengthens formatting features such as answer delimiters; their strict accuracy, which counts an answer only in the
trained format, rises far more than their lenient accuracy. \citet{mimo} observe that their model's first RL steps
mainly learn to adapt to the answer extraction function. \citet{calibfloor} split the accuracy change of
self-revision into a content part and format parts, where revision makes an answer readable or unreadable.
Concurrently, \citet{reportingprefs} find, on their own RL runs (GRPO) and with human calibration, that most of
the in-domain gain a strict scorer measures is that scorer's accuracy rising toward a lenient scorer's. We measure the same kind of
split under three other conditions: on released checkpoints, as read by extractors copied from public code, and on
out-of-domain multiple choice.

\paragraph{The extractor as instrument.} \citet{xfinder} measure the accuracy of regular-expression extractors and
replace them with a trained one, and \citet{verifierfalse} find that rule-based verifiers miss equivalent answers
written in other formats. \citet{rightanswer} find that rule-based extraction of free-text multiple-choice answers
often underestimates models, while LLM-based extractors make systematic errors. \citet{promptsens} trace much prompt
sensitivity to heuristic scoring, including rigid matching of extracted answers, and \citet{answermatching} argue
that answer matching should replace multiple choice. \citet{vulnscore} find that changing the extractor moves a score
only where a response's reasoning and its final verdict line disagree or no such line exists. An extractor's
accuracy does not say how much of a particular gain it produced, which turns on how the two models differ in where
they write their answers.

\section{Limitations}
\label{sec:limits}

\paragraph{No published gain is decomposed.} The gains we split are those that harness extractors report on outputs
we generated from the checkpoints; the two prompts and the pairing of the box-first extractors with these benchmarks
are ours (Section~\ref{sec:setup}). The SimpleRL-Zoo paper reports MMLU, MMLU-STEM and GPQA-Diamond accuracy before
and after RL for its checkpoints and reads the gains as generalization \citep{simplerlzoo}. It describes its
evaluation as using its own prompt, sampling at temperature 1.0, and a script built on Qwen2.5-Math's evaluation
code, which reads letters in the text. We did not reproduce that evaluation, so the 82\% is not an estimate for its numbers.
Appendix~\ref{app:survey} sets the published accuracies beside our readings of the same pairs.

\paragraph{Extractors that read the box first.} Nearly all of the gain behind the 82\% is reported by extractors that
read \verb|\boxed{}| first. An extractor that also reads letters in the text reports less format gain, as
Qwen2.5-Math's does (Section~\ref{sec:disagree}). The share also depends on running that extractor as its repository
runs it: its extraction function alone, without the letter step, gives a pooled share of 85\%
(Appendix~\ref{app:extractors}).

\paragraph{One release family at 7B.} The four pairs come from the SimpleRL releases and rest on three 7B base
models, and the Qwen base models carry nearly all of the reported gain. Whether the share holds in other
model families is open.

\paragraph{Prompts.} Both of our prompts keep the chat template's default system message, which for
Qwen2.5-Math-7B itself asks for \verb|\boxed{}|. A pilot that replaced it with its own request for the letter in
\verb|\boxed{}| gave a share of 31\% for Qwen2.5-Math-7B with SimpleRL-Zoo on MMLU, against 88\% and 86\% under our
boxed-instruction and plain prompts. Under the pilot's prompt the base model stated no answer that the fixed extractor
could read in 45\% of its outputs, so most of the gain that the fixed extractor reads there was a gain in stating an
answer at all (Appendix~\ref{app:omega}). The share is thus a property of the pair, the prompt and the extractors
together.

\paragraph{Scale.} The 82\% does not hold on the one larger pair with gains to split. On Qwen2.5-32B trained with
DAPO \citep{dapo}, which we set apart from the four pairs in advance, the format share is 2.1\% over 9 comparisons
(95\% CI $-18.5$\% to 19.9\%): the fixed extractor finds almost all of the gain that the harness extractors report.
Before generating these outputs we declared that a share below 40\% on this pair would make the result a property of
models of 7B and below. The pair also shows little of what the mechanism of Section~\ref{sec:position} needs, a
difference in where the two models write their answers: asked for the box, the 32B base model already boxes nearly
every answer, and without that request the DAPO model rarely boxes its own (Appendix~\ref{app:beyond}). One pair
cannot say whether this comes from scale, the DAPO recipe or the base model. Appendix~\ref{app:beyond} also gives a Mistral-Small-24B pair, which has too few
comparisons that meet both criteria to give a share.

\paragraph{The fixed extractor can fail.} When RL moves answers away from the places the fixed extractor reads, the
extractors can disagree even on the sign of the change. This happened in a pilot on a Llama-3.1-8B pair \citep{llama3},
where the fixed extractor reports a loss of 19 points. The extraction-rate criterion keeps that pilot's comparisons
out of the 82\% (Appendix~\ref{app:beyond}).

\paragraph{The readers.} The independent check of the fixed extractor used language models; no person has read the
responses.

\section{Conclusion}

When an RLVR model is credited with an out-of-domain gain, part of the credit can belong to the program that read its
answers. On four released 7B SimpleRL pairs, 82\% of the out-of-domain multiple-choice gain that harness extractors
report is the RL model writing its answer where those extractors look, in \verb|\boxed{}|, while the base model states
its answers in the text. On the one larger pair with gains to split, a 32B model, almost none of the reported gain is
format gain: its base model boxes its answers when asked, and its RL model rarely does when not. Where the releases we surveyed say how, their
out-of-domain multiple-choice results were almost always read out of generated text, so a format gain can arise in
them in the same way. We have not measured its size there, and an extractor that also reads letters in the text
reports less of it. Three practices make a format gain visible, and none needs RL to be re-run. Name the extractor
behind every reported number. Report the harness extractor's extraction rate on the base model: it can be measured
before any RL model exists, and in our main result the extractor that reads the most of the base models' outputs
reports the least format gain (Appendix~\ref{app:tracks}). And re-score the same outputs with a second extractor that reads both models' stated
answers, which measures the format gain itself.

\subsection*{Reproducibility}

Generation and scoring are separate, and everything after generation runs on a laptop. The ancillary files of this
arXiv version hold the analysis code, the frozen items with their hashes, the result files and the records behind the
appendices' counts, the independent readers' responses, and the SHA-256 hash of each of the 264 generation files.
Eight of those files ship with them; the other 256, which together exceed arXiv's size limit, are available from the
author (Appendix~\ref{app:cost}). Each generation records the hash of the item file it was built from, so the
evaluation set provably did not change between the two models.

\subsection*{AI use statement}

In this work, we used generative AI tools to help propose and refine hypotheses, to give feedback on the methodology
and experiments, and to help interpret results. We have not used generative AI tools to generate synthetic data sets
or for translation: the responses that the extractors and the readers read are outputs of publicly released
checkpoints, the object of this study. Qualitative or thematic data analysis is not applicable to this work. We also
used generative AI tools to edit prose, create figures, search the literature and the released checkpoints, and run
the experiment pipeline. We have reviewed all AI-assisted work: the study's design and structure, the decisions
recorded in its analysis files, the review and revision of all content and the verification of every result remain
with us, and every reported number was re-derived from the raw result files. We take responsibility for the final
content of this work, including text, claims or artifacts produced with the aid of generative AI. The study itself
also uses language models: the independent readers of Section~\ref{sec:readers} are language models by design.

\bibliographystyle{iclr2027_conference}
\bibliography{refs}

\appendix
\clearpage

In the appendices, the \emph{main result} is the 82\% of Section~\ref{sec:main}, pooled over its 46
comparisons; the \emph{grid} is everything Table~\ref{tab:grid} lists; and the \emph{letter step} is the step that
Qwen2.5-Math's evaluation loop runs after the extraction function: when the item's key is a letter from A to E and
the function's answer is not such a letter, it reads a letter from the text (Table~\ref{tab:app-harness} gives the
rule). The step is part of how that extractor is run
(Section~\ref{sec:setup} and Appendix~\ref{app:extractors}), and \emph{the extraction function alone} is the
extractor without it. Shares are given as fractions, so that
0.822 is the 82.2\% of the main text;
Tables~\ref{tab:robust} and~\ref{tab:strata}, which serve the body, keep percentages.

\section{How the population narrows}
\label{app:population}

This appendix gives the grid in counts (Table~\ref{tab:app-counts}), traces the 46 comparisons behind the main
result back to the 192 generation runs of the grid (Table~\ref{tab:app-chain}), and lists the release candidates that the admission rule of Section~\ref{sec:setup}
turned away. A \emph{combination} is one base--RL pair on one benchmark under one prompt; a comparison is a
combination read by one harness extractor.

\begin{table}[!htbp]
\caption{\textbf{The grid in counts.} Table~\ref{tab:grid} names what each row counts. A comparison is one pair,
benchmark, prompt and harness extractor.}
\label{tab:app-counts}
\centering
\small
\begin{tabular}{p{0.46\textwidth}p{0.46\textwidth}}
\toprule
base--RL pairs & 7, on 5 base models, from 3 releases \\
benchmarks & 8: 4 out of domain, 4 in domain; up to 500 items each \\
prompts & 2: boxed-instruction and plain \\
harness extractors & 9, from 6 repositories; 4 read multiple choice, 8 free-form answers \\
comparisons & 616 \\
\midrule
out-of-domain multiple choice, readable & 100 comparisons \\
and with a clearly positive reported gain: the main result & 46 comparisons, from 4 pairs \\
\bottomrule
\end{tabular}
\end{table}

\begin{table}[!htbp]
\caption{\textbf{From generation runs to the 46 comparisons of the main result.} Each row follows from the one
above it; the text explains the second row. The four groups are those of Table~\ref{tab:strata}.}
\label{tab:app-chain}
\centering
\small
\begin{tabular}{p{0.30\textwidth}rp{0.53\textwidth}}
\toprule
step & count & how it follows from the row above \\
\midrule
generation runs & 192 & 12 checkpoints $\times$ 8 benchmarks $\times$ 2 prompts \\
combinations & 112 & 7 base--RL pairs $\times$ 8 benchmarks $\times$ 2 prompts \\
comparisons & 616 & 70 multiple-choice combinations $\times$ 4 harness extractors $+$ 42 free-form
combinations $\times$ 8 \\
extraction-rate criterion met & 240 & out of domain: multiple choice 100, free-form 32; in domain: multiple
choice 28, free-form 80 \\
significant-gain criterion also met & 93 & out of domain: multiple choice 46, free-form 2; in domain: multiple
choice 10, free-form 35 \\
main result & 46 & out-of-domain multiple choice: 4 pairs, 3 benchmarks, 2 prompts, 20 combinations \\
\bottomrule
\end{tabular}
\end{table}

Of the first three rows, which build the population, only the second does not follow from the row above by
multiplication: it pairs each RL model with its base model. The 12 checkpoints are 5 base models and 7 RL models, and two of the base models have two RL
successors each:
Qwen2.5-Math-7B (SimpleRL-Zoo and SimpleRL-Zero) and Qwen2.5-1.5B (SimpleRL-Zoo and Open-Reasoner-Zero). The
outputs of these two base models therefore enter two pairs each, so the 12 checkpoints form 7 pairs, and each
pair has 16 combinations of benchmark and prompt. Multiple-choice combinations are read by the 4 harness
extractors that read letters and free-form ones by 8 (Appendix~\ref{app:extractors}), so each pair contributes
88 comparisons.

By pair, the comparisons that meet the extraction-rate criterion number 80 for Qwen2.5-Math-7B with
SimpleRL-Zero, 68 for Qwen2.5-Math-7B with SimpleRL-Zoo, 52 for Qwen2.5-7B, 28 for Mistral-7B-v0.1 and 12 for
DeepSeek-Math-7B. Neither pair on Qwen2.5-1.5B has any, because that base model runs to the token limit on
almost every item (Appendix~\ref{app:tracks}). Open-Reasoner-Zero-1.5B's 88 comparisons are among the 616, but
the pair is kept out of every pooled share: its base and RL models ship different chat templates, so the two would
not receive the same input. Figure~\ref{fig:position} is drawn over the other six pairs. On the four pairs of the
main result alone, 29.5\% of the base models' out-of-domain multiple-choice outputs that $\omega$ reads contain a
\verb|\boxed{}|, against 66.9\% of the RL models'. DeepSeek-Math-7B's 4 readable
out-of-domain multiple-choice comparisons, all on MMLU under the plain prompt, show no significant reported
gain, and its 2 comparisons that meet both criteria are in domain.

The 46 comparisons of the main result come from 20 combinations. By pair, 15 each come from Qwen2.5-Math-7B
with SimpleRL-Zoo and with SimpleRL-Zero, 14 from Qwen2.5-7B with SimpleRL-Zoo and 2 from
Mistral-7B-v0.1 with SimpleRL-Zoo; the three Qwen pairs carry 98.5\% of the summed reported gain. By benchmark,
17 are on ARC-Challenge, 16 on MMLU-Pro and 13 on MMLU. By prompt, 23 use the boxed-instruction prompt and 23
the plain prompt. By harness extractor, 18 are read by the MATH extractor, 17 by Open-Reasoner-Zero's, 9 by
Qwen2.5-Math's and 2 by simple-evals' multiple-choice extractor. All 46 have a positive format gain, and in 43
its 95\% interval excludes zero.

Candidates were screened in August 2026 against the admission rule of Section~\ref{sec:setup}: an RL model
trained by RL with verifiable rewards directly from a named public base model. Twelve did not meet it, almost
all because RL started from a checkpoint that was already distilled, instruction-tuned or fine-tuned:
Eurus-2-7B-PRIME (from a supervised fine-tuned model); five Skywork-OR1 models and AceReason-Nemotron-7B and
-14B (from DeepSeek-R1-distilled models); Light-R1-7B-DS (distilled, then fine-tuned); Light-R1-32B (from an
instruction-tuned model, trained with supervised fine-tuning and DPO); AceReason-Nemotron-1.1-7B (with a
supervised stage); and OpenRLHF's public checkpoints (RLHF or DPO after a supervised stage). A Llama-3.1-8B
pair from the SimpleRL-Zoo release met the rule, but its base model is access-gated, and it was run as a pilot.
Two larger pairs met the rule and were set apart from the pool before their generation;
Appendix~\ref{app:beyond} reports all three. The pool rests on one
release family for these reasons: the only pairs from other labs that met the rule, Open-Reasoner-Zero on
Qwen2.5-1.5B and DAPO on Qwen2.5-32B, are outside it, the first for its chat templates and the second as a
larger pair.

\section{The fixed extractor and its variants}
\label{app:omega}

This appendix states the rules of the fixed extractor $\omega$, compares it with three variants that read more
or less of each response, and shows how far a fixed extractor that reads the two models unequally moves the
format share.

\paragraph{Rules.} For a multiple-choice item, $\omega$ looks for six answer positions in this order, stops at
the first that occurs anywhere in the response, and takes its last occurrence:
\begin{enumerate}
\setlength{\itemsep}{0pt}
\item a letter alone inside \verb|\boxed{}|, as in \verb|\boxed{B}| or \verb|\boxed{(B)}|;
\item a letter inside \verb|\boxed{\text{}}|;
\item a letter after ``answer is'', ``the answer is'' or ``the correct answer is'', with or without a colon,
bold marks or parentheses;
\item a letter after ``Answer:'';
\item a line that holds only the letter, as ``B'', ``(B)'', ``B.'' or ``B)'';
\item a letter after ``option'' or ``choice''.
\end{enumerate}
Only the item's own option letters count, A to D for four options and up to A to J for ten, and a letter
followed directly by another letter is not read, so ``the answer is Because'' yields nothing. Positions 3, 4 and
6 ignore case. If no position occurs, $\omega$ returns nothing and the item is scored wrong for that model. For
a free-form item, $\omega$ reads the content of the last \verb|\boxed{}|, with nested braces balanced and an
enclosing \verb|\text{}| removed; failing that, the text after the last ``the answer is'' or ``final answer
is'', up to the next period or line end; failing that, the text after the last ``answer:''; otherwise nothing.
Free-form answers are compared with the key after a light normalisation (dollar signs, spaces, a final period,
a percent sign and thousands separators removed, and numbers compared as numbers). The comparison is the same
for every extractor and both models, except that the Qwen2.5-Math extractor also normalises the key as its
repository does (Appendix~\ref{app:extractors}).

\paragraph{Variants.} Three variants change how much $\omega$ reads (Table~\ref{tab:app-variants}). The strict
variant reads only the last \verb|\boxed{}|. It applies the same rule as the boxed-only harness extractor it is
compared with, so its format gain is zero by construction, and it admits no comparison: on every one it reads
under half of one model's outputs or reads the two models' at rates more than 20 points apart. The permissive
variant keeps $\omega$ and, where $\omega$ finds nothing, adds looser forms of the same cues that do not
require the letter to stand alone (``answer is'', ``answer:'' or ``answer ='' followed by a letter) and a letter
at the very end of the response; for free-form items it adds the last number in the response. The maximal
variant adds, after those, the last option letter anywhere in the response, so it almost never abstains. On the
18 comparisons that the three versions other than the strict one all admit, all from the three Qwen pairs, the
maximal variant
exceeds $\omega$ by 0.0184 (95\% CI 0.0004 to 0.0357, from a paired bootstrap of 10,000 draws over items).
Within this family the share therefore moves by less than 0.02 and $\omega$ gives the lowest value, although the
interval's lower end is too close to zero to fix the direction on its own. On its own comparisons the maximal
variant reads lower than $\omega$ only because the 5 comparisons it adds have a summed format gain of $-8.8$
points. All four versions were written for this study, so the table shows that the share is insensitive within
one family of rules, not that the family is right; Appendices~\ref{app:readers} and~\ref{app:robust} check
$\omega$ against readers and an extractor from outside it.

\begin{table}[!htbp]
\caption{\textbf{Reading more of each response moves the format share little, and upward on the same
comparisons.} Each version admits comparisons under its own extraction-rate criterion (own comparisons, with
that version's extraction rate on them); the last column holds the 18 comparisons that the other three versions
all admit. The harness extractor is the boxed-only rule, applied as in the strict variant; the
transcribed MATH extractor gives 0.854 on its own 18 comparisons of the main result
(Table~\ref{tab:robust}).}
\label{tab:app-variants}
\centering
\small
\begin{tabular}{lrcll}
\toprule
 & own & extraction rate & format share, own & format share, \\
fixed extractor & comparisons & base / RL (\%) & comparisons [95\% CI] & 18 common \\
\midrule
strict (\texttt{\textbackslash boxed\{\}} only) & 0 & --- & --- & --- \\
$\omega$ (abstaining) & 18 & 87.7 / 88.9 & 0.855 [0.814, 0.898] & 0.8551 \\
permissive & 18 & 88.4 / 89.0 & 0.858 [0.816, 0.901] & 0.8579 \\
maximal & 23 & 98.0 / 98.8 & 0.843 [0.797, 0.890] & 0.8735 \\
\bottomrule
\end{tabular}
\end{table}

\paragraph{A fixed extractor that reads the two models unequally.} An extractor whose extraction rate differs
between the two models moves the share far more than these variants do (Table~\ref{tab:app-pilot}). A pilot on
Qwen2.5-Math-7B and its SimpleRL-Zoo successor on MMLU had two fixed extractors of its own. Its permissive extractor
falls back on loose patterns, such as ``answer is'' or ``='' followed by any token. Its abstaining extractor reads
five of $\omega$'s six answer positions, all but \verb|\boxed{\text{}}|, and nothing else. On the grid's outputs for
that pair and benchmark under the boxed-instruction prompt (500 items), the permissive extractor gives a format
share of 0.5535, where $\omega$ gives 0.8805. It reads 80.4\% of the base model's outputs and 99.8\% of the RL
model's, a 19.4-point gap that the extraction-rate criterion admits, and so credits the RL model with a
fixed-extractor gain of 14.2 points where $\omega$ finds 3.8. It is not the permissive variant of
Table~\ref{tab:app-variants}: on these outputs it reads less of the base model's outputs than $\omega$ does. On the
pilot's own outputs it read an answer into responses that state none in 10.0\% of the base model's outputs and
4.2\% of the RL model's. Which model a guessing extractor favours depends on the outputs, so $\omega$ abstains
instead.

\paragraph{The pilot's own result.} The pilot generated its own outputs for the same pair and benchmark: 500 MMLU
items drawn separately from the grid's, at most 1,024 new tokens, and one prompt, which asks for the letter inside
\verb|\boxed{}| in a system message. That message replaces the default system message of the checkpoint's chat
template, which both prompts of the grid keep. On those outputs the MATH extractor reports a gain of 18.0 points
and $\omega$ reads 12.4, a format share of 0.311 (95\% CI 0.084 to 0.536), against 0.881 and 0.864 for the same
pair and benchmark under the grid's two prompts. Under the pilot's prompt the base model states no answer that
$\omega$ can read in 45.2\% of its outputs (24.2\% for the RL model); where $\omega$ reads an answer, the two
models' accuracies are 55.8\% and 56.7\%. Most of the RL model's gain under $\omega$ is then a gain in stating an
answer at all, which the format share does not count as format. $\omega$ reads 54.8\% of the base model's outputs
and 75.8\% of the RL model's, a gap of 21.0 points, so this comparison misses the extraction-rate criterion by one
point and would not enter the main result. We set that criterion after the pilot, for free-form comparisons in
which $\omega$ reads one model and not the other, and it is applied without regard to the sign of the format gain.
The share is thus a property of the pair, the prompt and the two extractors together: under three prompts, this
pair and benchmark give 0.31, 0.88 and 0.86.

\begin{table}[!htbp]
\caption{\textbf{A fixed extractor that reads more of the RL model's outputs lowers the format share.} The
grid's outputs for Qwen2.5-Math-7B and its SimpleRL-Zoo successor on MMLU under the boxed-instruction prompt
(500 items), read by a boxed-only harness extractor that reports a gain of 31.8 points.}
\label{tab:app-pilot}
\centering
\small
\begin{tabular}{lccc}
\toprule
fixed extractor & extraction rate, base / RL (\%) & fixed-extractor gain & format share \\
\midrule
the pilot's permissive extractor & 80.4 / 99.8 & 14.2 & 0.5535 \\
the pilot's abstaining extractor & 93.4 / 95.4 & 3.4 & 0.8931 \\
$\omega$ & 93.4 / 96.0 & 3.8 & 0.8805 \\
\bottomrule
\end{tabular}
\end{table}

\paragraph{Abstention inside the main result.} Over the 46 comparisons of the main result, $\omega$ abstains on
13.1\% of the base models' items and 12.0\% of the RL models' on average, a difference of $-1.06$ points, or
$-1.81$ weighted by reported gain. In 17 comparisons, from 9 combinations, the two abstention rates differ by
more than a quarter of the reported gain, in both directions (from $-9.8$ to $+11.4$ points, RL minus base).
The comparison of Figure~\ref{fig:example} is the most negative of them: $\omega$ finds no answer in 56 of the base
model's 500 outputs and in 7 of the RL model's. On 43 of the 56, Qwen2.5-Math's extractor reads the key, and these
43 are the 8.6 points by which it scores the base model above $\omega$, 87.0\% against 78.4\%
(Section~\ref{sec:disagree}). They state the
answer in forms that $\omega$ does not read: an option written out on a line of its own after a line ending in a
colon (40 of them), or ``boxed\{B\}'' without its backslash (27). On this comparison's 80 sampled items the
independent readers find the two models equally accurate, 90.0\% each, where $\omega$ reads a gain of 7.5 points
on the same responses. On these outputs, then, $\omega$'s gain of 9.8 points overstates the gain in answers, and
the share of 0.68 understates the format gain. No
comparison is dropped for this. Where $\omega$ abstains more on the RL model, it can produce a share above one
(Appendix~\ref{app:robust}), and Appendix~\ref{app:readers} measures what its abstentions cost each model.

\section{Independent readers}
\label{app:readers}

This appendix gives the design of the check in Section~\ref{sec:readers} as it was declared before the sample
was drawn, the readers' instructions, and the full results.

\paragraph{Population and sample.} The check was declared on the 50 comparisons that meet both criteria when
Qwen2.5-Math's extractor is scored with the extraction function alone. They come from 20 combinations and rest on
34 generation runs: 14 of base models and 20 of RL models. The 46 comparisons of the main result are among these
50 and come from the same 20 combinations, so the same sampled responses cover both. For each of the 14
base-model runs, 80 items were
drawn uniformly without replacement (seed 20260930) from the items present in that run and in every RL-model
run that shares it, and the same items were read in all of these runs. The sample holds 2,720 responses: 1,120
from base models and 1,600 from RL models.

\paragraph{Blinding and readers.} A reader saw the question, its lettered options and one response. It never
saw the key, which model wrote the response, or any extractor's reading. Responses carried random identifiers
and were shuffled across all 34 runs before being cut into batches of ten. Two readers read every response,
Claude Sonnet~5 (\texttt{claude-sonnet-5}) and Claude Haiku~4.5 (\texttt{claude-haiku-4-5-20251001}). Where
they differed, Claude Opus~5.5 (\texttt{claude-opus-5-5}) read the response with the same instructions, and its
reading decided. Each batch was a fresh call of the Claude command-line client in print mode
(\texttt{claude -p}), started in an empty temporary folder with session persistence off, so that a reader
saw nothing but its own batch. A response not returned in a readable form after
three attempts was to be recorded as missing.

\paragraph{Statistic and declared outcomes.} The readers' format share keeps each comparison's reported gain
on all items and corrects $\omega$'s fixed-extractor gain by the sample estimate, on the comparison's
combination, of how much the readers' reading changes the gain:
\begin{equation*}
g_{\mathrm{readers}} = \gfix - \delta, \qquad
\delta = \left(\mathrm{acc}_{\omega} - \mathrm{acc}_{\mathrm{readers}}\right)_{\mathrm{RL}}
       - \left(\mathrm{acc}_{\omega} - \mathrm{acc}_{\mathrm{readers}}\right)_{\mathrm{base}},
\end{equation*}
with accuracies in points on the combination's sampled items. The share is then pooled over the 50 comparisons
as in Section~\ref{sec:method}, and the statistic is its difference $\Delta$ from $\omega$'s share. The 95\%
interval of $\Delta$ comes from 10,000 bootstrap draws of the sampled items within each base-model run, the same
draw applied to that run and to every RL-model run that shares it (seed 20260930). Three outcomes were declared
before the run: the readers agree with $\omega$ if $|\Delta| \le 0.05$ and the interval lies inside
$[-0.10, +0.10]$; they move the share if $|\Delta| > 0.05$ and the interval excludes zero; every other result
is inconclusive. Whatever the outcome, the main result stays $\omega$'s. The declaration also fixed the
statistics of Tables~\ref{tab:app-readers-group} and~\ref{tab:app-readers-class}, except the rows marked
descriptive. Its SHA-256 is written into the program that ran the readers, which refuses to run if the
declaration has changed.

\paragraph{Instructions.} Every reader received the following instructions, shown here with lines broken for
print, then the batch's items, each with its identifier, question, options and the response between markers,
and last the line ``Now return one JSON line per item, in the order given, and nothing else.''
{\footnotesize
\begin{verbatim}
You are checking what answer a language model gave to multiple-choice
questions. Below are several items. Each item has an id, a question,
its lettered options, and one response written by a model.

For each item, report the option letter that the response commits to
as its final answer to the question shown, or NONE.

Rules:
1. Do not solve the question and do not judge whether the answer is
   right. Report what the response says.
2. If the response gives an answer and later changes it, report the
   last answer it gives to this question.
3. If the response goes on to write and answer other questions, ignore
   them. Report only its answer to the question shown.
4. If the response names an option by its text rather than by its
   letter, report that option's letter.
5. If the response commits to no option, or to more than one option,
   report NONE.

Return exactly one line of JSON per item, in the order given, and
nothing else:
{"id": "<id>", "final": "<one letter or NONE>", "quote": "<the
shortest phrase from the response that shows this answer, or an empty
string>"}
\end{verbatim}
}

\begin{table}[!htbp]
\caption{\textbf{Where $\omega$ reads a letter, the readers read the same one; its misses are abstentions.}
Percentages of the responses read in each group, except in the descriptive rows, whose denominator is the count above
them. The descriptive rows were computed after the declared statistics had been read. The two groups are not
matched one to one (a sampled Qwen2.5-Math-7B base response is compared with two RL models), so the difference
between their accuracies is not a gain.}
\label{tab:app-readers-group}
\centering
\small
\begin{tabular}{p{0.62\textwidth}rr}
\toprule
 & base models & RL models \\
\midrule
responses read & 1,120 & 1,600 \\
$\omega$ and the readers agree (the same letter, or neither finds one) & 88.3 & 88.8 \\
$\omega$ finds no answer, the readers find one & 10.5 & 9.9 \\
$\omega$ reads a letter, the readers find none & 0.6 & 0.3 \\
$\omega$ and the readers read different letters & 0.5 & 0.9 \\
accuracy under $\omega$ & 53.7 & 55.1 \\
accuracy under the readers & 60.0 & 62.3 \\
\midrule
\multicolumn{3}{l}{\emph{descriptive}} \\
responses in which $\omega$ reads a letter & 941 & 1,391 \\
\quad the readers read the same letter (\%) & 98.6 & 98.6 \\
responses in which $\omega$ finds no answer & 179 & 209 \\
\quad the readers find a letter (\%) & 65.9 & 76.1 \\
\quad the readers' letter is the key (\%) & 39.1 & 52.2 \\
\bottomrule
\end{tabular}
\end{table}

\begin{table}[!htbp]
\caption{\textbf{Agreement with the readers by the answer position $\omega$ read.} Agreement is the percentage
of responses on which the readers' reading equals $\omega$'s: the same letter, or, in the last row, no answer.
Positions follow the order of Appendix~\ref{app:omega}. Rates over fewer than 20 responses are in parentheses.}
\label{tab:app-readers-class}
\centering
\small
\begin{tabular}{lrrrr}
\toprule
 & \multicolumn{2}{c}{base models} & \multicolumn{2}{c}{RL models} \\
\cmidrule(lr){2-3}\cmidrule(lr){4-5}
answer position read by $\omega$ & responses & agreement (\%) & responses & agreement (\%) \\
\midrule
\texttt{\textbackslash boxed\{X\}} & 292 & 99.3 & 1,036 & 99.9 \\
\texttt{\textbackslash boxed\{\textbackslash text\{X\}\}} & 1 & (100.0) & 35 & 100.0 \\
``answer is X'' & 491 & 98.8 & 247 & 96.8 \\
``Answer: X'' & 147 & 98.6 & 50 & 84.0 \\
a line holding only the letter & 2 & (0.0) & 1 & (0.0) \\
``option X'' or ``choice X'' & 8 & (87.5) & 22 & 90.9 \\
no answer found & 179 & 34.1 & 209 & 23.9 \\
\bottomrule
\end{tabular}
\end{table}

\paragraph{Result.} Under the readers' reading the format share over the 50 comparisons of the declaration is
0.801 (95\% CI 0.735 to 0.869), against
0.847 under $\omega$: $\Delta = -0.0465$ (4.7 points), with an interval from $-0.1125$ to $+0.0213$ (its lower end
11.2 points below zero). The readers' interval is
$\omega$'s share plus the 95\% interval of the difference between the two, from the bootstrap over the 2,720 responses
they read with each comparison's full-sample gains held fixed; unlike the main result's interval, it resamples
only the sampled items, not the items behind each comparison's gains. The point lies inside
$\pm0.05$, but the interval's lower end passes $-0.10$ and the interval contains zero, so the declared outcome
is inconclusive. The same statistic over the 46 comparisons of the main result, computed after the readings were
known, puts the readers' share at 0.771 (95\% CI 0.696 to 0.849), against 0.822 under $\omega$:
$\Delta = -0.0508$, with an interval from $-0.1259$ to $+0.0266$. Its point lies just outside $\pm0.05$ and its
interval contains zero, so the declared rule would read it as inconclusive too. The first two readers gave the
same reading on 98.2\% of the 2,719 responses that both
returned (Cohen's $\kappa$ 0.978). The tie-break reader read the 51 responses on which they differed or one of
them returned no reading, and no response is missing after it.

Where $\omega$ reads a letter, the readers read the same letter on 98.6\% of responses in both groups
(Table~\ref{tab:app-readers-group}). Most of $\omega$'s disagreements with the readers are abstentions, and
where it abstains, the readers' letter is the key more often for the RL models (52.2\%) than for the base models
(39.1\%). The abstentions therefore cost the RL models slightly more accuracy than the base models, which is
why the readers' share is lower than $\omega$'s. By answer position (Table~\ref{tab:app-readers-class}), the
readers agree with $\omega$ on at least 96.8\% of the responses in which $\omega$ reads \verb|\boxed{}| or ``the
answer is''. Among the positions read in at least 20 responses, they agree least often where $\omega$ reads
``Answer:'' in an RL model's response (84.0\% of 50); on the three responses in which $\omega$ reads a line holding
only the letter, they never agree. The release also holds a blind
sample of 120 responses, 60 from base and 60 from RL models (seed 20260931), prepared for a human check. We have not
run that check, and no result here depends on it.

\section{Robustness in full}
\label{app:robust}

This appendix gives every check of Section~\ref{sec:robust} in full. Each check changes one thing and keeps
the rest of the main result: which comparisons enter the ratio (Tables~\ref{tab:app-denominator}
to~\ref{tab:app-threshold}), leaving out one pair, benchmark, extractor or prompt (Table~\ref{tab:app-loo}), the
harness extractor
(Table~\ref{tab:app-extractor-shares}), shares above one (Table~\ref{tab:app-above-one}), and an external fixed
extractor (Table~\ref{tab:app-external}).

\paragraph{Intervals.} Our 95\% intervals on the format share come from a bootstrap that resamples items within
each benchmark and applies the same draw to every comparison built on that benchmark, because every extractor reads
the same outputs. They measure the uncertainty from item sampling only. Variation between pairs and between
benchmarks, of which the main result has four and three, is shown by leaving each one out
(Table~\ref{tab:app-loo}).

\begin{table}[!htbp]
\caption{\textbf{The values of Figure~\ref{fig:robust}.} One thing changes per row; everything else
is as in the main result (first row). The three rows on the harness extractors one by one and on Qwen2.5-Math's
extraction function alone are not drawn in the figure. Rows that admit more comparisons can include pairs outside the main result's
four, from the six of Figure~\ref{fig:position}. Shares in percent; changes are computed before rounding. The readers'
interval is built differently from the first row's, from the sampled responses they read rather than from the
benchmarks' resampled items
(Appendix~\ref{app:readers}). $^{\ast}$On the 18 comparisons of
the three Qwen pairs that all three versions admit, with the \texttt{\textbackslash boxed\{\}} rule that MATH and
Open-Reasoner-Zero share as the harness extractor.}
\label{tab:robust}
\centering
\small
\setlength{\tabcolsep}{4pt}
\begin{tabular}{p{0.52\textwidth}rl}
\toprule
what changes & comparisons & format share (\%) \\
\midrule
main result: both criteria, both prompts, fixed extractor & 46 & \textbf{82.2} [77.4, 87.2] \\
boxed-instruction prompt only & 23 & 78.8 \\
plain prompt only & 23 & 85.6 \\
every readable comparison & 100 & 79.3 \\
any non-zero reported gain & 96 & 75.0 \\
extraction-rate criterion, five settings & 39--50 & 79.3--82.2 \\
harness extractor, own comparisons: MATH, Open-Reasoner-Zero, Qwen2.5-Math & 18, 17, 9 & 85.4, 83.5, 56.2 \\
the same three, on the comparisons all three admit & 8 & 86.3, 84.0, 58.0 \\
Qwen2.5-Math's extraction function alone: all extractors, Qwen2.5-Math's own & 50, 13 & 84.7, 84.2 \\
fixed extractor: ours, permissive (looser patterns), maximal (any option letter)$^{\ast}$ & 18 & 85.5,
85.8, 87.4 \\
leave out one pair, benchmark, or harness extractor (largest change, points) & & $\le$3.2, $\le$1.6, $\le$3.1 \\
comparisons with a share above 100\% (a negative fixed-extractor gain) removed & 40 & 79.5 \\
independent LLM readers in place of the fixed extractor & 46 & 77.1 [69.6, 84.9] \\
\bottomrule
\end{tabular}
\end{table}

\paragraph{Which comparisons enter the ratio.} The three rules for which comparisons enter the ratio move the
share by at most 0.073, and both other rules lower it (Table~\ref{tab:app-denominator}); under the same rules each
of the other three domain and answer-form groups moves by 0.515 to 5.434 (Appendix~\ref{app:strata}). The 4 comparisons with a reported
gain of exactly zero add nothing to the summed reported gain but 32.0 points to the summed format gain, which
separates the second rule from the third. The 54 comparisons that the second rule adds to the main result's 46
include 37 with a negative reported gain, 20 of them read by simple-evals' extractor. Summed over all 54 added
comparisons, the reported gain is $-280.6$ points and the format gain $-253.0$, a ratio of 0.902. That ratio is
above the main result's 0.822, so admitting them takes away a larger part of the summed format gain than of the
summed reported gain, and the share falls.

\begin{table}[!htbp]
\caption{\textbf{Admitting comparisons without a significant reported gain lowers the share.} Out-of-domain
multiple choice, all comparisons that meet the extraction-rate criterion.}
\label{tab:app-denominator}
\centering
\small
\begin{tabular}{p{0.30\textwidth}rrrl}
\toprule
 & & \multicolumn{2}{c}{summed gain (points)} & \\
\cmidrule(lr){3-4}
comparisons admitted & comparisons & format & reported & format share [95\% CI] \\
\midrule
significant reported gain (main result) & 46 & 849.2 & 1032.8 & 0.8222 [0.7735, 0.8716] \\
every readable comparison & 100 & 596.2 & 752.2 & 0.7926 [0.6712, 0.9249] \\
any non-zero reported gain & 96 & 564.2 & 752.2 & 0.7501 [0.6345, 0.8745] \\
\bottomrule
\end{tabular}
\end{table}

Five settings of the extraction-rate criterion move the share within 0.7933 to 0.8222, and the setting used,
at least 50\% of each model's outputs read with the two rates within 20 points, gives the highest of the five
(Table~\ref{tab:app-settings}). Without any criterion, four comparisons enter, all under the boxed-instruction
prompt. Three are of DeepSeek-Math-7B: MMLU and ARC-Challenge read by Open-Reasoner-Zero's extractor, and
ARC-Challenge read by Qwen2.5-Math's. On these two benchmarks $\omega$ reads 55.2\% and 51.6\% of the base model's
outputs, against 77.6\% and 85.6\% of the RL model's. The fourth is of Qwen2.5-1.5B with SimpleRL-Zoo, on MMLU-Pro
read by Qwen2.5-Math's extractor, where $\omega$ reads 3.6\% of the base model's outputs and 10.8\% of the RL
model's. Dropping the chat-template
rule of Appendix~\ref{app:population} as well readmits Open-Reasoner-Zero-1.5B, whose base model's outputs
$\omega$ reads at 3.6--9.2\% against 81.8--98.0\% for the RL model. Its fixed-extractor gain then rests on a
base-model accuracy read from under a tenth of the outputs, so its format gain is not identified; with it included,
the setting without any criterion reads 0.3564 over 68 comparisons.

\begin{table}[!htbp]
\caption{\textbf{The extraction-rate criterion used gives the highest share of five settings.} Out-of-domain
multiple choice, comparisons with a significant reported gain.}
\label{tab:app-settings}
\centering
\small
\begin{tabular}{ccrrl}
\toprule
minimum rate, each model (\%) & maximum gap (points) & comparisons & pairs & format share \\
\midrule
none & none & 50 & 6 & 0.7976 \\
30 & 30 & 47 & 5 & 0.8191 \\
50 & 20 & 46 & 4 & 0.8222 (used) \\
60 & 15 & 45 & 4 & 0.8191 \\
70 & 10 & 39 & 3 & 0.7933 \\
\bottomrule
\end{tabular}
\end{table}

Because the ratio weights each comparison by its reported gain, comparisons near the significant-gain
threshold carry little weight: those with a reported gain under 5 points hold 0.4\% of the summed reported gain,
and those under 10 points 8.8\%. Dropping the comparisons below 5, 10, 15 or 20 points keeps the share within
0.0455 of 0.8222, and not in one direction (Table~\ref{tab:app-threshold}).

\begin{table}[!htbp]
\caption{\textbf{Comparisons with small reported gains carry little weight.} The main result with the
comparisons below each reported gain removed.}
\label{tab:app-threshold}
\centering
\small
\begin{tabular}{rrrrl}
\toprule
reported gain below & comparisons & their share of the & comparisons & format share \\
(points) & removed & summed reported gain (\%) & kept & \\
\midrule
5 & 1 & 0.4 & 45 & 0.8191 \\
10 & 11 & 8.8 & 35 & 0.8387 \\
15 & 18 & 17.0 & 28 & 0.8495 \\
20 & 25 & 27.5 & 21 & 0.8677 \\
\bottomrule
\end{tabular}
\end{table}

\paragraph{Leaving out one pair, benchmark, extractor or prompt.} Leaving out any one pair moves the share by at
most 0.032, any one benchmark by at most 0.016, and any one harness extractor by at most 0.031
(Table~\ref{tab:app-loo}); these bounds are the largest changes rounded up. Leaving out one harness extractor never
removes the last-\verb|\boxed{}| rule, which the MATH and Open-Reasoner-Zero extractors share; leaving out both
leaves Qwen2.5-Math's 9 comparisons and simple-evals' 2, with a share of 0.634. Leaving out either Qwen2.5-Math-7B pair does not leave out that base model,
since its SimpleRL-Zoo and SimpleRL-Zero pairs share its outputs. Leaving out one prompt moves the share most, to 0.7879 or
0.8557, because the two prompts' shares differ (Section~\ref{sec:robust}) and each carries about half of the
summed reported gain, the plain prompt 50.6\%. Under the boxed-instruction prompt, fewer of the base models'
outputs that $\omega$ reads contain a \verb|\boxed{}| than of the RL models', 45.4\% against 69.5\%, over the pairs of
Figure~\ref{fig:position}.

\begin{table}[!htbp]
\caption{\textbf{Leaving out any one pair, benchmark or harness extractor moves the share by at most 0.032.}
The main result (0.8222) without each group. Zoo and Zero are the SimpleRL-Zoo and SimpleRL-Zero RL models.}
\label{tab:app-loo}
\centering
\small
\begin{tabular}{lll}
\toprule
group & left out & format share \\
\midrule
pair & Qwen2.5-Math-7B, Zoo & 0.8345 \\
 & Qwen2.5-Math-7B, Zero & 0.8410 \\
 & Qwen2.5-7B, Zoo & 0.7907 \\
 & Mistral-7B-v0.1, Zoo & 0.8182 \\
\midrule
benchmark & MMLU & 0.8137 \\
 & MMLU-Pro & 0.8203 \\
 & ARC-Challenge & 0.8376 \\
\midrule
harness extractor & MATH (\texttt{\textbackslash boxed\{\}} only) & 0.7920 \\
 & Open-Reasoner-Zero & 0.8137 \\
 & Qwen2.5-Math & 0.8493 \\
 & simple-evals, multiple choice & 0.8182 \\
\midrule
prompt & plain (boxed-instruction prompt only) & 0.7879 \\
 & boxed-instruction (plain prompt only) & 0.8557 \\
\bottomrule
\end{tabular}
\end{table}

\paragraph{The harness extractor.} The three extractors that carry the result do not agree
(Table~\ref{tab:app-extractor-shares}). The two that read only the last \verb|\boxed{}| give 0.854 and 0.835 on
their own comparisons, and Qwen2.5-Math's extractor, which also reads letters in the text, gives 0.562 on
its own. The 8 comparisons that all three admit show the same gap. The two that agree run the same algorithm
(Appendix~\ref{app:extractors}), so theirs is not the agreement of two independent rules. Together they carry
89.1\% of the summed reported gain of the main result, and Qwen2.5-Math's extractor carries 9.4\%. Eight of its 9
comparisons are on ARC-Challenge and MMLU-Pro, which its repository does not evaluate; there it applies its
free-form rules, which read the last \verb|\boxed{}| before any ``the answer is'' (Table~\ref{tab:app-harness}), and
then the letter step. Its one comparison
on MMLU, the pairing that its repository also makes, gives 0.378. Extractors that read \verb|\boxed{}| first thus
carry 97.6\% of the summed reported gain.

Run as its extraction function alone, without the letter step, Qwen2.5-Math's extractor gives 0.842 over 13
comparisons, against 0.562 over 9 as its repository runs it. The three extractors then give 0.829 to 0.854 on the
12 comparisons that all three admit. No comparison is admitted for all four extractors, because simple-evals'
extractor meets the significant-gain criterion in only 2 of its 25 readable comparisons.

\begin{table}[!htbp]
\caption{\textbf{Qwen2.5-Math's extractor, which also reads letters in the text, reports less format gain
than the two that read only the box; its extraction function alone reports about as much.} Out-of-domain multiple
choice. Each extractor has 25 comparisons that meet the
extraction-rate criterion; the significant-gain criterion removes some (removed) and keeps the rest (own
comparisons). Weight is the extractor's part of the summed reported gain of the main result.}
\label{tab:app-extractor-shares}
\centering
\small
\setlength{\tabcolsep}{4.5pt}
\begin{tabular}{lrrrll}
\toprule
 & & own & weight & \multicolumn{2}{c}{format share} \\
\cmidrule(lr){5-6}
harness extractor & removed & comparisons & (\%) & own comparisons & 8 common \\
\midrule
MATH (\texttt{\textbackslash boxed\{\}} only) & 7 & 18 & 49.0 & 0.8537 & 0.8632 \\
Open-Reasoner-Zero & 8 & 17 & 40.1 & 0.8349 & 0.8400 \\
Qwen2.5-Math & 16 & 9 & 9.4 & 0.5617 & 0.5805 \\
\quad on ARC-Challenge & & 5 & & 0.569 & \\
\quad on MMLU-Pro & & 3 & & 0.610 & \\
\quad on MMLU & & 1 & 0.9 & 0.378 & \\
simple-evals, multiple choice & 23 & 2 & 1.5 & 1.0909 & --- \\
\midrule
Qwen2.5-Math, function alone & 12 & 13 & & 0.842 & --- \\
\bottomrule
\end{tabular}
\end{table}

\paragraph{Shares above one.} Six of the 46 comparisons have a share above one, because $\omega$ reads a
small negative gain where the harness extractor reads a positive one (Table~\ref{tab:app-above-one}). They carry
10.9\% of the summed format gain and 7.8\% of the summed reported gain, and without them the share is 0.7947.
Five of the six come from two MMLU-Pro combinations under the plain prompt, in which $\omega$ reads fewer of the
RL model's outputs than of the base model's: 66.8\% against 76.4\% for Qwen2.5-Math-7B with SimpleRL-Zoo, and
69.0\% against 80.4\% for Qwen2.5-7B. There the fixed-extractor gains are $-2.4$ and $-1.6$ points, with
intervals that include zero. The six pass both criteria and are kept: dropping comparisons by their share
would select on the outcome.

\begin{table}[!htbp]
\caption{\textbf{The six comparisons with a share above one.} Gains in accuracy points, grouped by
combination.}
\label{tab:app-above-one}
\centering
\small
\begin{tabular}{lrrr}
\toprule
harness extractor & reported gain & fixed-extractor gain & format share \\
\midrule
\multicolumn{4}{l}{\emph{Qwen2.5-7B with SimpleRL-Zoo, MMLU-Pro, plain prompt}} \\
MATH (\texttt{\textbackslash boxed\{\}} only) & 25.2 & $-1.6$ & 1.063 \\
Open-Reasoner-Zero & 23.2 & $-1.6$ & 1.069 \\
Qwen2.5-Math & 7.6 & $-1.6$ & 1.211 \\
\multicolumn{4}{l}{\emph{Qwen2.5-Math-7B with SimpleRL-Zoo, MMLU-Pro, plain prompt}} \\
MATH (\texttt{\textbackslash boxed\{\}} only) & 11.4 & $-2.4$ & 1.211 \\
Open-Reasoner-Zero & 8.4 & $-2.4$ & 1.286 \\
\multicolumn{4}{l}{\emph{Mistral-7B-v0.1 with SimpleRL-Zoo, MMLU-Pro, boxed-instruction prompt}} \\
simple-evals, multiple choice & 4.4 & $-2.4$ & 1.545 \\
\bottomrule
\end{tabular}
\end{table}

\paragraph{An external fixed extractor.} Qwen2.5-Math's extractor, which this study did not write (the authors of
the Qwen2.5-Math checkpoints ship it, in evaluation code adapted from math-evaluation-harness), gives 0.775 in
$\omega$'s place over its own 25 comparisons, against $\omega$'s 0.855 over its 18 (Table~\ref{tab:app-external}).
The two sets share 18 comparisons, all of $\omega$'s, and on those the pooled shares are within 0.03 of each other,
with an interval that includes zero. That agreement is a cancellation. On the 6 MMLU comparisons, where the
external extractor applies its letter rule, it gives 1.053 against $\omega$'s 0.858; on the 12 MMLU-Pro and
ARC-Challenge comparisons, where it applies its free-form rules and then the letter step, it gives 0.718 against
0.854. Comparison by comparison the two shares differ by 0.25 on average, and in 5 of the 18 the two extractors
disagree on the sign of the RL model's gain.

Item by item, over every out-of-domain multiple-choice
output of the six grid pairs other than Open-Reasoner-Zero (36 combinations), the two give the same score on
85.1\% of the base models' outputs and 86.1\% of the RL models'. $\omega$ scores the base models 0.91 points higher
than the Qwen2.5-Math extractor does and the RL models 0.76 points higher, so item by item the two extractors
differ by about as much on either model. The check does not show, however, that an extractor written outside this study
reproduces $\omega$'s share comparison by comparison. The check of $\omega$ that remains is the independent readers
of Appendix~\ref{app:readers}, who read the same letter as $\omega$ in 98.6\% of the responses in which it reads
one.

\begin{table}[!htbp]
\caption{\textbf{With an external extractor in $\omega$'s place, the pooled share on the comparisons both admit is
within 0.03 of $\omega$'s, by offsetting differences.} Out-of-domain multiple choice, the six grid pairs other than
Open-Reasoner-Zero, with the boxed-only harness extractor as in Table~\ref{tab:app-variants}; $\omega$ admits
comparisons of the three Qwen pairs only. Each difference is Qwen2.5-Math's share minus $\omega$'s, with an
interval paired over items (10,000 bootstrap draws). The last two blocks split the 18 comparisons that both admit:
on MMLU the external extractor applies its letter rule, on the other two benchmarks its free-form rules and then
the letter step.}
\label{tab:app-external}
\centering
\small
\begin{tabular}{llrcl}
\toprule
 & & & extraction rate & \\
comparisons & fixed extractor & $n$ & base / RL (\%) & format share [95\% CI] \\
\midrule
own & $\omega$ & 18 & 87.7 / 88.9 & 0.855 [0.814, 0.898] \\
own & Qwen2.5-Math & 25 & 94.3 / 95.9 & 0.775 [0.728, 0.824] \\
both admit & $\omega$ & 18 & & 0.8551 \\
both admit & Qwen2.5-Math & 18 & & 0.8277 \\
both admit & difference & 18 & & $-0.0274$ [$-0.0640$, $+0.0093$] \\
\midrule
MMLU & $\omega$ & 6 & & 0.858 \\
MMLU & Qwen2.5-Math & 6 & & 1.053 \\
MMLU & difference & 6 & & $+0.195$ [$+0.125$, $+0.263$] \\
\midrule
other two & $\omega$ & 12 & & 0.854 \\
other two & Qwen2.5-Math & 12 & & 0.718 \\
other two & difference & 12 & & $-0.136$ [$-0.180$, $-0.094$] \\
\bottomrule
\end{tabular}
\end{table}

\section{The harness extractors}
\label{app:extractors}

This appendix lists the nine harness extractors (Table~\ref{tab:app-harness}), what each returns when it finds no
answer, how they were copied, where the executable versions depart from the source and by how much
(Table~\ref{tab:app-departures}), and how multiple choice is scored.

\begin{table}[!htbp]
\caption{\textbf{The nine harness extractors.} Each is copied from its repository at the commit shown; files are
given relative to the repository root. ``Nothing'' means that the extractor returns no value.}
\label{tab:app-harness}
\centering
\scriptsize
\begin{tabular}{p{0.15\textwidth}p{0.29\textwidth}p{0.30\textwidth}p{0.13\textwidth}}
\toprule
\raggedright extractor & \raggedright source and role & \raggedright what it reads & \raggedright when nothing
matches \tabularnewline
\midrule
\multicolumn{4}{l}{\emph{multiple choice and free-form}} \tabularnewline
\raggedright MATH (\texttt{\textbackslash boxed\{\}} only) & \raggedright \nolinkurl{hendrycks/math} @ 940d1bf
(\nolinkurl{modeling/dataset/util.py}, \nolinkurl{modeling/eval_math_gpt.py}); the MATH benchmark's evaluation
code & \raggedright the last \texttt{\textbackslash boxed\{\}} &
\raggedright nothing \tabularnewline
\raggedright Qwen2.5-Math & \raggedright \nolinkurl{QwenLM/Qwen2.5-Math} @ cb098e1
(\nolinkurl{evaluation/parser.py}, \nolinkurl{evaluation/math_eval.py}); evaluation code, with a copy in the
SimpleRL-Zoo code; the extraction function, then the letter step of the loop that calls it & \raggedright
benchmark-aware: for AQuA-RAT, SAT-Math and MMLU (run as MMLU-STEM), the letter rule: a letter A--E standing
alone, the first after the last ``answer is'' or ``choice is'' (only in the first paragraph when either occurs
twice), else the last in the response, read case-insensitively; otherwise, in turn, ``final answer is \$X\$. I
hope'', the last \texttt{\textbackslash boxed\{\}}, the last ``the answer is'', the last ``final answer is'' (or its
Chinese form), the last number; on MMLU-Pro the
result is narrowed to the last letter A--E in it; then normalised. Then, on an item keyed A--E whose result is not
such a letter, the letter step applies the letter rule to the whole response; this can change an answer only on
ARC-Challenge and on the 287 of MMLU-Pro's 500 items keyed A--E & \raggedright on items keyed A--E, the
response's own text when it holds no letter; otherwise the empty string
\tabularnewline
\raggedright Open-Reasoner-Zero & \raggedright \nolinkurl{Open-Reasoner-Zero/Open-Reasoner-Zero} @ f6d1ec7
(\nolinkurl{orz/ppo/tools/math_utils.py}); RL training code, which applies it only to a
\texttt{\textbackslash boxed\{\}} inside \texttt{<answer>} tags; we apply it to the whole response & \raggedright
the last \texttt{\textbackslash boxed\{\}} & \raggedright the whole response \tabularnewline
\midrule
\multicolumn{4}{l}{\emph{multiple choice only}} \tabularnewline
\raggedright simple-evals, multiple choice & \raggedright \nolinkurl{openai/simple-evals} @ 5e623c2
(\nolinkurl{common.py}, \nolinkurl{gpqa_eval.py}); evaluation harness & \raggedright the first ``Answer: X'',
with X one of A--D & \raggedright nothing \tabularnewline
\midrule
\multicolumn{4}{l}{\emph{free-form only}} \tabularnewline
\raggedright simple-evals, math & \raggedright \nolinkurl{openai/simple-evals} @ 5e623c2
(\nolinkurl{common.py}, \nolinkurl{math_eval.py}); evaluation harness & \raggedright the text after the first
``Answer:'', to the line end & \raggedright nothing \tabularnewline
\raggedright lm-evaluation-harness GSM8K, strict-match & \raggedright
\nolinkurl{EleutherAI/lm-evaluation-harness} @ b733598 (\nolinkurl{lm_eval/tasks/gsm8k/gsm8k.yaml},
\nolinkurl{lm_eval/filters/extraction.py}); evaluation harness & \raggedright the first ``\#\#\#\# number'' &
\raggedright the string ``[invalid]'' \tabularnewline
\raggedright lm-evaluation-harness GSM8K, flexible-extract & \raggedright as above & \raggedright the last run
of two or more digits, dollar signs, points or commas, or else of digits & \raggedright ``[invalid]'', or a run
of punctuation with no digit \tabularnewline
\raggedright verl GSM8K, strict & \raggedright verl's reward function, as shipped in
\nolinkurl{hkust-nlp/simpleRL-reason} @ 2204dd1 (\nolinkurl{verl/utils/reward_score/gsm8k.py}); RL training
reward & \raggedright the first ``\#\#\#\# number'' & \raggedright nothing \tabularnewline
\raggedright verl GSM8K, flexible & \raggedright as above & \raggedright the last run of digits, points or
commas that is not a bare point & \raggedright nothing, or ``.'' when every run is a bare point
\tabularnewline
\bottomrule
\end{tabular}
\end{table}

\paragraph{When nothing matches.} Given a response that holds no answer they can read, the nine extractors do
five different things. lm-evaluation-harness's two GSM8K rules return the string ``[invalid]'' on most such
inputs; Qwen2.5-Math's returns the empty string except on items keyed A--E, where it returns any letter in the
response or else the response's text; simple-evals' two extractors, MATH's and verl's strict rule
return nothing; the two flexible rules can return a run of punctuation with no digit, because their number
patterns admit one; and Open-Reasoner-Zero's \verb|\boxed{}| reader, applied to a whole response as we apply it, returns that
response. For lm-evaluation-harness's flexible
rule, responses with no digit return ``[invalid]'' 99.02\% of the time and a punctuation run 0.98\% of the time,
over 76,428 responses from 11 checkpoints. In this study's scoring, a returned nothing is an abstention and is
scored wrong; a returned string, empty or not, is an answer, and it is scored wrong unless it matches the key.

\paragraph{lm-evaluation-harness's default MMLU.} The task that \texttt{lm\_eval --tasks mmlu} runs has no
extraction step: it scores the log-likelihood of each of the four options and takes the most likely, so it
never reads generated text and has no format gain to produce. Its generative variant,
\texttt{mmlu\_generative}, has to be selected by name. The harness's default MMLU-Pro and BBH tasks, by
contrast, generate text and extract the answer from it. This holds at the pinned commit and on the harness's main
branch as read in August 2026.

\paragraph{Copying.} Each extractor was transcribed verbatim from the files of its repository fetched at the
pinned commit. The record of these copies holds the nine extractors with their helper functions and task
configurations; its 11 entries that hold source text match the same files fetched again at the same commits. The
executable versions used for scoring were written from these copies and tested against them, and every departure
from the source is listed with its measured effect in Table~\ref{tab:app-departures}. The source of an extractor is
its function together with the step of its evaluation script that reads the answer. For Qwen2.5-Math that step is
the letter step of its evaluation loop, and its executable version includes it, so its extractor is run as its
repository runs it; the first row of the table gives the effect of leaving the step out. Open-Reasoner-Zero's
training code applies its function only to a \verb|\boxed{}| inside \texttt{<answer>} tags, and we apply it to the
whole response (Table~\ref{tab:app-harness}).

\begin{table}[!htbp]
\caption{\textbf{Every departure of the executable extractors from their source, with its measured effect.}}
\label{tab:app-departures}
\centering
\footnotesize
\begin{tabular}{p{0.20\textwidth}p{0.40\textwidth}p{0.31\textwidth}}
\toprule
\raggedright extractor & \raggedright departure & \raggedright measured effect \tabularnewline
\midrule
\raggedright Qwen2.5-Math, letter step & \raggedright none in score: the step is applied as the repository's
evaluation loop applies it (\nolinkurl{evaluation/math_eval.py}, lines 358--371; in the SimpleRL-Zoo code,
\nolinkurl{hkust-nlp/simpleRL-reason} @ 2204dd1, the same fourteen lines are lines 402--415 of
\nolinkurl{examples/simplelr_math_eval/math_eval.py}). Before the step the loop also strips the response and
normalises the extracted answer again; these two operations are left out, and they change no multiple-choice
score &
\raggedright leaving it out, as the extraction function alone does, gives a pooled share of 0.847 over 50
comparisons, against the main result's 0.822 over 46 (and 0.811 with the step on the same 50); Qwen2.5-Math's own
share of 0.842 over 13, against 0.562 over 9, with a reported gain larger by 4.8 to 59.0 points on 12 of the 13;
and $+12.8$ on the outputs of Figure~\ref{fig:example}, against $+1.0$. Tables~\ref{tab:disagree}
and~\ref{tab:app-larger} give the rest
\tabularnewline
\raggedright Qwen2.5-Math, other rules & \raggedright none: the source's rules are applied (option letters A--E,
the per-benchmark setting, upper-casing of the response) & \raggedright departing from these rules in the
extraction function moves 12 of 24 multiple-choice comparisons by at least 1.6 points, up to 18.8
\tabularnewline
\raggedright Qwen2.5-Math, key & \raggedright the key is normalised with the function applied to the answer, as
the source does & \raggedright 381 free-form items change, all to correct; no multiple-choice key changes
\tabularnewline
\raggedright verl GSM8K, flexible & \raggedright reproduced on purpose: when every candidate run is a bare point,
the source returns it rather than nothing & \raggedright fires on 2.853\% of the base models' and 1.833\% of the
RL models' free-form outputs, mostly on BBH \tabularnewline
\raggedright simple-evals, math & \raggedright the captured answer is stripped of surrounding spaces &
\raggedright 5 of 33,000 outputs change \tabularnewline
\raggedright simple-evals, multiple choice & \raggedright the captured letter is upper-cased & \raggedright 0 of
43,428 outputs change \tabularnewline
\raggedright MATH and Open-Reasoner-Zero & \raggedright a shared helper also accepts
\texttt{\textbackslash boxed} followed by a space, which neither source does & \raggedright fires on 0 of
111,168 responses \tabularnewline
\raggedright Open-Reasoner-Zero, use & \raggedright its training code applies the function only to a
\texttt{\textbackslash boxed\{\}} inside \texttt{<answer>} tags; we apply it to the whole response &
\raggedright with no \texttt{\textbackslash boxed\{\}}, the whole response is returned and scored on its first
character, which credits 6.99\% of the base models' and 4.61\% of the RL models' out-of-domain multiple-choice
outputs as correct; this lowers the share, by an amount not computed. Used inside \texttt{<answer>} tags, as its
code uses it, it would read no answer here: none of the 45,000 out-of-domain multiple-choice outputs of the other 15
checkpoints contains such a tag \tabularnewline
\raggedright lm-evaluation-harness GSM8K, both rules & \raggedright the source's stripping of spaces from the
match is omitted & \raggedright none possible: neither pattern admits a space \tabularnewline
\raggedright verl GSM8K, strict & \raggedright none: commas and dollar signs are removed, as in the source &
\raggedright none \tabularnewline
\bottomrule
\end{tabular}
\end{table}

\paragraph{First-character scoring.} Multiple choice is scored on the first character of the extracted
answer, compared with the key, for every extractor and both models. This matters for Open-Reasoner-Zero's
extractor, which returns the whole response when it finds no \verb|\boxed{}|: a response that opens with
``Answer'', for instance, is then scored as option A. Over the 33,000 out-of-domain multiple-choice outputs of
the 11 checkpoints whose pair receives the same input (every grid checkpoint except Open-Reasoner-Zero-1.5B), this
extractor returns a whole response longer than 40 characters on 78.76\% of the base models' outputs and 49.67\% of the RL models', and credits a correct answer this way on 6.99\% and 4.61\%. The extra credit falls more on the base models, which
lowers the reported gain and therefore the format share; its size on the main result is not computed. Under
exact matching, the MATH and Open-Reasoner-Zero extractors give the same correctness on 100.00\% of items, and
under first-character scoring on 94.31\%. The two therefore run one algorithm, and the gap between their shares
in Table~\ref{tab:app-extractor-shares} comes from the scoring rule, not from extraction.

\paragraph{The extractors on the same outputs.} Table~\ref{tab:disagree} gives the values of
Figure~\ref{fig:extractors}. Without simple-evals' two extractors, which look for an ``Answer:'' line that neither
of our prompts asks for and that simple-evals' own prompt does, the median spread between harness extractors is
9.6 points. The table's indented rows give the values with Qwen2.5-Math's extraction function alone.

\begin{table}[!htbp]
\caption{\textbf{Different extractors report different gains on the same outputs.} Top: on the outputs of Figure~\ref{fig:example}
(Qwen2.5-7B and SimpleRL-Zoo, ARC-Challenge, boxed-instruction prompt, 500 items), the gain each multiple-choice
harness extractor reports and the fixed extractor's reading.
Middle: over the grid's 112 combinations of pair, benchmark and prompt, the spread between harness extractors
(largest minus smallest reported gain), and the readable out-of-domain comparisons in which simple-evals scores the
RL model at exactly 0\%. Bottom: the 17 out-of-domain multiple-choice results reported in 18 recent RLVR and
reasoning-model releases, by how they were produced (Appendix~\ref{app:survey}).}
\label{tab:disagree}
\centering
\small
\begin{tabular}{p{0.62\textwidth}r}
\toprule
\multicolumn{2}{l}{\emph{the outputs of Figure~\ref{fig:example}: reported gain by extractor (points)}} \\
\midrule
simple-evals, multiple choice (``Answer: X'') & $-1.8$ \\
Qwen2.5-Math & $+1.0$ \\
\quad the extraction function alone & $+12.8$ \\
MATH (\texttt{\textbackslash boxed\{\}} only) & $+30.6$ \\
Open-Reasoner-Zero & $+30.6$ \\
fixed extractor $\omega$ & $+9.8$ \\
\midrule
\multicolumn{2}{l}{\emph{the grid}} \\
harness extractors' spread, points: median (quartiles) & 15.1 (6.7, 33.0) \\
\quad largest & 78.8 \\
\quad median without simple-evals' two extractors & 9.6 \\
\quad with Qwen2.5-Math's extraction function alone: median, median without simple-evals' two, largest & 15.4,
7.8, 94.4 \\
readable out-of-domain comparisons with the RL model at exactly 0\% under simple-evals & 14 of 25 \\
\midrule
\multicolumn{2}{l}{\emph{17 results from 18 releases, by how they were produced}} \\
extracting the answer from generated text & 13 of 17 \\
log-likelihood scoring & 1 of 17 \\
not stated & 3 of 17 \\
\bottomrule
\end{tabular}
\end{table}

\section{The other groups}
\label{app:strata}

This appendix gives the three groups of Table~\ref{tab:strata} other than out-of-domain multiple choice: how
each moves under the rules for which comparisons enter the ratio (Table~\ref{tab:app-strata}), why the domain
contrast within multiple choice is unresolved, and what the in-domain free-form gains consist of
(Table~\ref{tab:app-idff}).

\begin{table}[!htbp]
\caption{\textbf{Reported and fixed-extractor gains by domain and answer form.} Mean gains per comparison, in
accuracy points, over the comparisons that pass both criteria of Section~\ref{sec:criteria}, with the fixed
extractor's extraction rate on the base and RL models' outputs. The format share is printed only where it is stable.
For each of the other three groups, the shares under the rule of Section~\ref{sec:criteria} and under the two
alternatives of Appendix~\ref{app:robust} lie more than 50 points apart (Table~\ref{tab:app-strata}).}
\label{tab:strata}
\centering
\small
\setlength{\tabcolsep}{4.5pt}
\begin{tabular}{lrrrrlc}
\toprule
 & & \multicolumn{2}{c}{mean gain (points)} & format & & extraction rate (\%) \\
\cmidrule(lr){3-4}
 & comparisons & reported & fixed extr. & share (\%) & 95\% CI & base / RL \\
\midrule
\multicolumn{7}{l}{\emph{out of domain}} \\
\quad multiple choice & 46 & 22.5 & 4.0 & \textbf{82.2} & [77.4, 87.2] & 86.9 / 88.0 \\
\quad free-form       & 2  & 6.4  & $-2.2$ & --- & & 78.0 / 92.8 \\
\multicolumn{7}{l}{\emph{in domain}} \\
\quad multiple choice & 10 & 18.5 & 5.3 & --- & & 72.0 / 73.1 \\
\quad free-form       & 35 & 8.7  & 9.9 & --- & & 95.3 / 98.6 \\
\bottomrule
\end{tabular}
\end{table}

\begin{table}[!htbp]
\caption{\textbf{Across the three rules, the share of out-of-domain multiple choice moves by 0.07, that of each
other group by more than 0.5.} Format share by
group under each rule for which comparisons enter the ratio, with the number of comparisons in parentheses, and
the spread across the three rules. The first row is the main result.}
\label{tab:app-strata}
\centering
\small
\setlength{\tabcolsep}{4pt}
\begin{tabular}{lllll}
\toprule
 & significant reported & every readable & any non-zero & \\
group & gain [95\% CI] & comparison & reported gain & spread \\
\midrule
\multicolumn{5}{l}{\emph{out of domain}} \\
\quad multiple choice & 0.8222 (46) [0.7735, 0.8716] & 0.7926 (100) & 0.7501 (96) & 0.0721 \\
\quad free-form & 1.3438 (2), not quoted & 6.7778 (32) & 5.6389 (21) & 5.434 \\
\multicolumn{5}{l}{\emph{in domain}} \\
\quad multiple choice & 0.7150 (10) [0.5736, 0.8867] & 0.1996 (28) & 0.2308 (26) & 0.5154 \\
\quad free-form & $-0.1435$ (35) [$-0.208$, $-0.086$] & $-0.8701$ (80) & $-0.1688$ (50) & 0.7266 \\
\bottomrule
\end{tabular}
\end{table}

\paragraph{Out of domain, free-form.} Free-form answers out of domain rest on BBH alone. Of its 32 comparisons
that meet the extraction-rate criterion, 2 meet the significant-gain criterion, both from one combination
(Qwen2.5-Math-7B with SimpleRL-Zero, boxed-instruction prompt), whose reported gain of 6.4 points $\omega$ reads
as $-2.2$ while reading 78.0\% of the base model's outputs and 92.8\% of the RL model's. Two comparisons from
one combination with a 14.8-point gap in extraction rate cannot carry a share, so whether the format gain
reaches free-form answers out of domain is untested.

\paragraph{In domain, multiple choice.} This group has 10 comparisons. Its share moves from 0.369 to 0.829 when
one pair is left out and from 0.465 to 0.771 when one benchmark is left out, a movement of up to 0.346 that is
wider than its own interval.

\paragraph{The domain contrast is unresolved.} Within multiple choice, out-of-domain minus in-domain is
$+0.107$ (95\% CI $-0.057$ to 0.271) under the significant-gain criterion. Under the other two rules the in-domain
share falls to 0.200 and 0.231, the contrast becomes 0.593 and 0.519, and the two groups' intervals no longer
overlap: 0.7926 [0.6712, 0.9249] against 0.1996 [$-0.0469$, 0.4917] for every readable comparison, 0.7501
[0.6345, 0.8745] against 0.2308 [0.0021, 0.5033] for any non-zero reported gain. Which way the contrast reads is
thus set by the rule, a consequence of 10 in-domain comparisons against 46, and the data cannot say whether the
format gain is specific to out-of-domain evaluation.

\paragraph{In domain, free-form.} The share of $-0.144$ is a mixture of extractors (Table~\ref{tab:app-idff}).
The two boxed extractors read almost exactly what $\omega$ reads (a share of $+0.013$). Three extractors report
smaller gains than $\omega$ does. The table gives the shortfall over each extractor's own reported gain; over
$\omega$'s gain it is 20.6\% (Qwen2.5-Math's), 23.5\% (lm-evaluation-harness's flexible GSM8K rule) and 26.8\% (verl's
flexible rule). The two flexible rules take the last number anywhere in
the response; Qwen2.5-Math's reads \verb|\boxed{}| and ``the answer is'' first and falls back on the last number
only when neither occurs. The other three free-form extractors, simple-evals' and the two strict GSM8K rules,
have no comparison that meets both criteria. The 35 comparisons come from the three Qwen pairs, on MATH-500 and
GSM8K.

\begin{table}[!htbp]
\caption{\textbf{The in-domain free-form share is a mixture of extractors.} The 35 in-domain free-form comparisons
that meet both criteria, by harness extractor; gains summed over comparisons, in accuracy points.}
\label{tab:app-idff}
\centering
\small
\begin{tabular}{lrrrr}
\toprule
 & & \multicolumn{2}{c}{summed gain (points)} & \\
\cmidrule(lr){3-4}
harness extractor & comparisons & reported & fixed-extractor & format share \\
\midrule
MATH (\texttt{\textbackslash boxed\{\}} only) & 8 & 77.8 & 76.8 & 0.013 \\
Open-Reasoner-Zero & 8 & 77.8 & 76.8 & 0.013 \\
Qwen2.5-Math & 8 & 61.0 & 76.8 & $-0.259$ \\
lm-evaluation-harness GSM8K, flexible & 4 & 36.4 & 47.6 & $-0.308$ \\
verl GSM8K, flexible & 7 & 50.8 & 69.4 & $-0.366$ \\
\bottomrule
\end{tabular}
\end{table}

The grid also holds contrasts larger than the main result whose intervals exclude zero: multiple choice minus
free-form in domain, 0.8585 [0.6906, 1.0264], and out-of-domain minus in-domain for free-form answers, 1.4873
[0.383, 2.5916]. Each rests on a group that moves by an order of magnitude more than the main result under the
same rules, so neither is quoted as a finding. Multiple choice minus free-form out of domain is $-0.5216$
[$-1.6519$, 0.6087].

\section{Beyond the grid: larger pairs and the Llama pilot}
\label{app:beyond}

This appendix reports three pairs outside the grid: Mistral-Small-24B with SimpleRL-Zoo and Qwen2.5-32B with
DAPO, generated separately, each model on two GPUs, and a 100-item pilot on Llama-3.1-8B
with SimpleRL-Zoo.

\paragraph{The larger pairs.} Both were generated with the grid's items, prompts, decoding (greedy,
\texttt{float16}, at most 2048 new tokens), harness extractors and $\omega$; only the number of GPUs per model
differs. Both were set apart from the pooled share before their generation, whatever they show, and a declaration
fixed then how each result would be read. A pair with fewer than 5
comparisons meeting both criteria has no gain to decompose. With 5 or more, a share of at least 0.60 counts as
the result holding and a share below 0.40 as it failing: for Mistral-Small-24B across family and scale
together, for Qwen2.5-32B across scale within the Qwen family. Between 0.40 and 0.60, the declaration reads
Mistral-Small-24B as indeterminate and declares no reading for Qwen2.5-32B. Its hash was written into the
generation job before the job ran. Table~\ref{tab:app-larger} gives the outcome.

\begin{table}[!htbp]
\caption{\textbf{Neither larger pair reproduces the main result.} Out-of-domain multiple choice. The last row
scores Qwen2.5-Math's extractor with its extraction function alone (Appendix~\ref{app:extractors}).}
\label{tab:app-larger}
\centering
\small
\setlength{\tabcolsep}{4pt}
\begin{tabular}{p{0.19\textwidth}rrrlp{0.21\textwidth}}
\toprule
 & & & both & format share & \\
pair & comparisons & readable & criteria & [95\% CI] & declared reading \\
\midrule
Mistral-Small-24B, SimpleRL-Zoo & 24 & 8 & 3 & 0.560 [0.370, 0.760] & no gain to decompose (fewer than 5) \\
Qwen2.5-32B, DAPO & 24 & 24 & 9 & 0.021 [$-0.185$, 0.199] & \raggedright collapses with scale (below 0.40) \tabularnewline
\quad\raggedright the extraction function alone & 24 & 24 & 10 & 0.436 [0.328, 0.538] & none declared (between 0.40 and 0.60) \\
\bottomrule
\end{tabular}
\end{table}

On Qwen2.5-32B almost none of the reported gain is format gain. The pair's 9 comparisons that meet both criteria
have a summed format gain of $+2.0$ points over a summed reported gain of 94.0, a share of 0.021
(Table~\ref{tab:app-larger}). Leaving out any one harness extractor, or taking either prompt alone, gives a share
between $-0.04$ and 0.10 (Table~\ref{tab:app-32b}), so the main result is not reproduced on this pair under any
single extractor's removal. Under the declaration, a share below 0.40 on this pair makes the main result a property
of models of 7B and below (Section~\ref{sec:limits}).

The pair also shows little of the difference in where the two models write their answers on which a format gain
rests (Section~\ref{sec:position}). Under the boxed-instruction prompt the 32B base model writes \verb|\boxed{}| in
99.2\% of its out-of-domain multiple-choice outputs, against 42.2\% for the base models of the four pairs of the main
result, and its two comparisons there report $+6.4$ points where $\omega$ reads $+5.8$. Under the plain prompt the
DAPO model writes \verb|\boxed{}| in 14.8\% of its outputs (the four pairs' RL models 54.8\%, its base model 1.1\%).
MATH's extractor then reads 0\% to 3\% of the base model's outputs and 12\% to 19\% of the RL model's, so the gains it
reports there, 9.8 to 12.2 points, come from the RL model's few boxed answers. $\omega$ reads most of both models'
outputs and finds gains of about the same size, 10.4 to 12.8 points. The two gains agree in size without resting on
the same answers. The pair differs from the grid's in scale, in RL recipe and in lab, and one pair cannot separate
the three.

With Qwen2.5-Math's extraction function alone, the pair has 10 comparisons that meet both criteria and a share of
0.436, between the two cut-offs, where nothing was declared (Table~\ref{tab:app-larger}). Two comparisons then
carry the share: the two read by Qwen2.5-Math's extractor have a summed format gain of $+82.4$ points, more than
the pair's total of $+79.2$ over a summed reported gain of 181.8. Without that extractor the share is $-0.042$;
without MATH's it is 0.555, and without Open-Reasoner-Zero's 0.568. The letter step changes both of these
comparisons (Appendix~\ref{app:extractors}): with it, one no longer meets the significant-gain criterion, and the
other's format gain is $+5.2$ points. Either way, the two extractors that read only the box show no format gain
on this pair (Table~\ref{tab:app-32b}), so the main result does not carry over to it.

On Mistral-Small-24B, $\omega$ meets the extraction-rate criterion in 8 of the 24 comparisons, and 3 of those
meet the significant-gain criterion, below the declared floor of 5, so the pair's share of 0.560 is reported
and not used. Two of the three are read by simple-evals' extractor (summed format gain $+28.2$ points) and one by
Open-Reasoner-Zero's ($-2.0$), all under the plain prompt. The declared reading for this outcome is that the
pair has no gain to decompose on our outputs, like DeepSeek-Math-7B in the grid; it is not evidence
about the share. The SimpleRL-Zoo paper reports MMLU gains for all three pairs under its own evaluation, which we
did not reproduce (Table~\ref{tab:app-published}).

\begin{table}[!htbp]
\caption{\textbf{On the 32B pair, almost none of the reported gain is format gain, whichever extractor is left
out.} Qwen2.5-32B with DAPO, the 9 comparisons that meet both criteria.}
\label{tab:app-32b}
\centering
\small
\begin{tabular}{lrrl}
\toprule
harness extractor & its comparisons & its summed format gain & format share without it \\
\midrule
MATH (\texttt{\textbackslash boxed\{\}} only) & 4 & $+0.6$ & 0.0260 \\
Open-Reasoner-Zero & 4 & $-3.8$ & 0.0997 \\
Qwen2.5-Math & 1 & $+5.2$ & $-0.0421$ \\
\midrule
prompt & comparisons & share of the weight (\%) & format share \\
\midrule
plain & 7 & 86 & 0.0099 \\
boxed-instruction & 2 & 14 & 0.0937 \\
\midrule
all & 9 & & 0.0213 \\
\bottomrule
\end{tabular}
\end{table}

\paragraph{The Llama pilot.} The Llama-3.1-8B pair of the SimpleRL-Zoo release met the admission rule, but its
base model's weights are access-gated; they were taken from a public mirror whose four weight files have the
same SHA-256 hashes as the gated repository's. The pilot generated 50 items each of ARC-Challenge, MMLU and
MMLU-Pro under the boxed-instruction prompt; Table~\ref{tab:app-llama} covers the first two. RL moved this pair's
answers away from the positions
$\omega$ reads, the opposite of every Qwen pair: $\omega$ reads 90\% of the base model's ARC-Challenge outputs
and 68\% of the RL model's, and 70\% and 38\% on MMLU, so both benchmarks fail the extraction-rate criterion,
with gaps of 22 and 32 points. MMLU-Pro is left out because $\omega$ reads only 6\% of its base model's outputs
(the pilot ran $\omega$ with the four letters A to D on this ten-option benchmark); with it, over 150 items,
$\omega$'s reading under that setting is $-13.3$ points (95\% CI $-21.3$ to $-5.3$). Neither model writes a
readable \verb|\boxed{}|, even when asked.

$\omega$, which scores a missing answer as wrong, puts the RL model 19 points below its base model. Conditional
on $\omega$ reading an item, the RL model is not worse: 75.5\% against 73.8\%, $+1.7$ points, over 53 and 80
items. The pilot therefore shows a collapse in extraction rate, not a loss of capability. On the same outputs the
MATH extractor reports exactly 0.0, because it reads neither model, and Open-Reasoner-Zero's reports $+5.0$,
through its whole-response fall-back and first-character scoring (Appendix~\ref{app:extractors}). A harness
extractor can thus report no change where $\omega$ reports a large one, and can produce a rise.

\begin{table}[!htbp]
\caption{\textbf{One pair, five readings.} Gain of Llama-3.1-8B's SimpleRL-Zoo successor over its base model on
100 items (50 each of ARC-Challenge and MMLU, boxed-instruction prompt), in accuracy points.}
\label{tab:app-llama}
\centering
\small
\begin{tabular}{lr}
\toprule
extractor & gain [95\% CI] \\
\midrule
$\omega$ (fixed extractor) & $-19.0$ [$-31$, $-8$] \\
MATH (\texttt{\textbackslash boxed\{\}} only) & 0.0 [0, 0] \\
simple-evals, multiple choice & $-19.0$ [$-30$, $-8$] \\
Qwen2.5-Math & $-16.0$ [$-28$, $-5$] \\
Open-Reasoner-Zero & $+5.0$ [1, 10] \\
\bottomrule
\end{tabular}
\end{table}

\section{What the share tracks}
\label{app:tracks}

This appendix gives what the share tracks across base models and RL recipes: the base model's boxing rate, its
fit across pairs, the same base model under two RL recipes, the change in boxing rate between the two models, and
the harness
extractors' extraction rates on the base models.

\paragraph{The base model's boxing rate.} A combination's boxing rate is the percentage of the base model's
outputs on it that contain \verb|\boxed{}|, and its share is pooled over its comparisons. Across the 20
combinations of the main result, the share falls as the boxing rate rises: $r = -0.434$ (95\% CI $-0.687$ to
$-0.082$), a slope of $-0.0042$ per percentage point. The rate runs from 1.0\% (Qwen2.5-7B, ARC-Challenge, plain
prompt) to 73.2\% (Qwen2.5-Math-7B, MMLU-Pro, boxed-instruction prompt), and nine combinations exceed 38\%. The
20 rates take only 14 distinct values, because the SimpleRL-Zoo and SimpleRL-Zero pairs are compared with the same
base-model outputs, so that 12 combinations share their rate two by two; the bootstrap treats the 20 as independent, so the interval is
narrower than the design supports. Over the 46 comparisons, where each rate repeats once per harness extractor,
the correlation is weaker: $r = -0.275$. The correlation is across combinations, so it does not rank base models.

\paragraph{Across pairs.} Averaged over each pair's comparisons, the base models box 6.0\% of their outputs for
Mistral-7B, 29.2\% for Qwen2.5-7B, and 35.9\% (SimpleRL-Zero) and 38.3\% (SimpleRL-Zoo) for Qwen2.5-Math-7B.
Qwen2.5-Math-7B's two values differ only because its two pairs do not have the same comparisons. The pairs' shares are
1.091, 0.867, 0.762 (SimpleRL-Zero) and 0.798 (SimpleRL-Zoo). Over these four points $r = -0.986$, but one of
them is Mistral-7B, whose share comes from 2 comparisons and exceeds one; without it $r = -0.829$, over three
points from two base models. The fit across pairs is therefore not evidence; only the weak correlation across
combinations above is.

\paragraph{The same base model, two RL recipes.} SimpleRL-Zoo and SimpleRL-Zero both start from Qwen2.5-Math-7B
and have 15 comparisons each in the main result, 13 of them matched: the same combination read by the same
harness extractor. On the 13, with base-model outputs, benchmark and extractor all fixed, the shares are 0.8146
for SimpleRL-Zoo and 0.7904 for SimpleRL-Zero, a gap of 0.024 that can come only from the RL model. On each pair's
own 15 comparisons the shares are 0.7975 and 0.7624, a gap of 0.035, which differs from the matched gap by two
unmatched comparisons on each side. For SimpleRL-Zoo these are MMLU under the plain prompt, read by
Open-Reasoner-Zero's extractor, and MMLU-Pro under the boxed-instruction prompt, read by Qwen2.5-Math's; for
SimpleRL-Zero, MMLU and MMLU-Pro under the plain prompt, both read by Qwen2.5-Math's. No interval is computed for
the gap, so it is not shown to exceed sampling noise, and as evidence of a recipe effect it is exploratory.

\paragraph{The change in boxing rate.} The share does not track how much more often the RL model boxes than its
base model (Table~\ref{tab:app-change}). Within out-of-domain multiple choice, over the 20 combinations with each
combination's share pooled as in the rest of this paper, the correlation between that change and the share is
$r = -0.03$. The other rows of Table~\ref{tab:app-change} average each combination's comparison shares instead, which
gives $-0.109$ (95\% CI $-0.546$ to 0.544). Controlling each for the other, with
pooled shares, the base model's rate keeps its association with the share (partial $r = -0.455$) and the change
does not ($-0.154$). A large positive correlation appears only across groups: over
the 93 comparisons of all four groups that meet both criteria, $r = 0.567$, and over their 35 combinations,
0.402. That is a contrast between groups. Within out-of-domain multiple choice the correlation is near zero
over combinations, and within the groups the sign is not uniform: it reverses in in-domain free-form only, and is
positive in in-domain multiple choice. No dose--response on the change is detectable within a group at this grid's
size.

\begin{table}[!htbp]
\caption{\textbf{The change in boxing rate tracks the share across groups, not within out-of-domain multiple
choice.} Correlation between the change
in the rate of \texttt{\textbackslash boxed\{\}} from base to RL model and the format share. Across comparisons,
each combination's change repeats once per harness extractor; across combinations, each combination's share is
the mean of its comparisons' shares, except in the row marked pooled, which sums gains as in the rest of the
paper.}
\label{tab:app-change}
\centering
\small
\begin{tabular}{lrrrl}
\toprule
 & \multicolumn{2}{c}{across comparisons} & \multicolumn{2}{c}{across combinations} \\
\cmidrule(lr){2-3}\cmidrule(lr){4-5}
group (both criteria met) & $n$ & $r$ & $n$ & $r$ \\
\midrule
out of domain, multiple choice & 46 & $+0.089$ & 20 & $-0.109$ [$-0.546$, 0.544] \\
\quad the same, pooled per combination & & & 20 & $-0.03$ \\
out of domain, free-form & 2 & --- & 1 & --- \\
in domain, multiple choice & 10 & $+0.404$ & 6 & $+0.21$ \\
in domain, free-form & 35 & $-0.574$ & 8 & $-0.87$ \\
all four groups & 93 & $+0.567$ & 35 & $+0.402$ \\
\bottomrule
\end{tabular}
\end{table}

\paragraph{Harness extractors' extraction rates on the base models.} Table~\ref{tab:app-matrix} gives, for each
base model and multiple-choice harness extractor, the percentage of the base model's out-of-domain
multiple-choice outputs from which the extractor returns an answer whose first character is an option letter. The
count also takes in an empty return, which three of the extractors can give (Qwen2.5-Math's when it finds nothing,
MATH's and Open-Reasoner-Zero's on an empty \verb|\boxed{}|) and which no key matches. The rates in parentheses leave
empty returns out, and 15 of the 20 are lower. Under the MATH extractor the rate runs from 3.9\% (Mistral-7B) to
32.0\% (Qwen2.5-Math-7B), or from 0.0\% (Qwen2.5-1.5B) to 30.5\% without empty returns. Under simple-evals'
extractor, whose rates are the same under both counts, the base models change order, with Mistral-7B first at
33.8\% and Qwen2.5-1.5B last at 1.6\%. Across the 20 entries, the extractor accounts for 78.5\% of the variance, the
base model for 9.4\%, and the residual for 12.1\%; without empty returns the three are 75.2\%, 14.1\% and 10.7\%.
Changing the extractor with the base model fixed moves the rate by a median of 72.7 points, and changing the base
model with the extractor fixed by a median of 30.15 points. The ratio of the two medians, 2.41, has a permutation
$p$-value below 0.0001: none of 20,000 shuffles of the 20 entries, with the row and column structure destroyed
(seed 20260818), reached it. Without empty returns the medians are 66.7 and 31.35 points, the ratio 2.13 and the
$p$-value 0.0003. Under either count the rate is a property of the pairing of base model and extractor, and it
varies more between extractors than between base models.

The extraction-rate criterion does not select base models with high rates. Mistral-7B, at 3.9\% under the MATH
extractor (3.6\% without empty returns), has all 24 of its out-of-domain multiple-choice comparisons admitted,
while the two pairs on Qwen2.5-1.5B, at 21.6\% (0.0\%), have none, because $\omega$ reads only 6.7\% of that base
model's outputs: it runs to the token limit on 98.7\% of items.

A harness extractor's extraction rate on the base model needs neither a fixed extractor nor an RL model, and in the
main result it goes with the share. Over the seven combinations of base model and extractor that contribute
comparisons to the main result, the share is lower where the rate is higher: $r = -0.858$ with pooled shares. The
seven are not independent, and the interval of that value, $-0.987$ to $-0.376$, treats them as if they were: four
are one rule, the last \verb|\boxed{}|, on two base models, and two are Qwen2.5-Math's extractor on two base models.
Without those two, $r = -0.461$ over five, so the relation rests mostly on that one extractor, which reads the most
of the base models' outputs and reports the least format gain. Leaving out any one combination, including the one
whose share exceeds one, gives $-0.934$ to $-0.765$. Without empty returns $r = -0.845$ ($-0.988$ to $-0.302$). A
mean of per-comparison shares would give $-0.783$ ($-0.987$ to $-0.321$). Seven points do not support a prediction
beyond these combinations.

\begin{table}[!htbp]
\caption{\textbf{A base model's extraction rate depends on the extractor more than on the model.} Percentage of
each base model's out-of-domain multiple-choice outputs (3,000 per base model, both prompts) from which the
harness extractor returns an answer whose first character is an option letter, with an empty return counted among
them; in parentheses, the percentage without empty returns. Open-Reasoner-Zero's rate includes
whole responses returned when it finds no \texttt{\textbackslash boxed\{\}} (Appendix~\ref{app:extractors}).}
\label{tab:app-matrix}
\centering
\small
\begin{tabular}{lrrrr}
\toprule
 & simple-evals, & MATH & & \\
base model & multiple choice & (\texttt{\textbackslash boxed\{\}} only) & Open-Reasoner-Zero & Qwen2.5-Math \\
\midrule
Qwen2.5-Math-7B & 26.0 & 32.0 (30.5) & 62.8 (61.3) & 93.1 (89.7) \\
Qwen2.5-7B & 5.7 & 31.4 (29.0) & 39.3 (36.9) & 91.9 (88.1) \\
Qwen2.5-1.5B & 1.6 & 21.6 (0.0) & 71.9 (50.3) & 74.3 (63.4) \\
DeepSeek-Math-7B & 4.6 & 7.4 (0.5) & 15.6 (8.0) & 71.4 (67.2) \\
Mistral-7B-v0.1 & 33.8 & 3.9 (3.6) & 66.2 (65.8) & 82.0 (76.2) \\
\bottomrule
\end{tabular}
\end{table}

\section{How released RLVR and reasoning-model checkpoints report out-of-domain multiple choice}
\label{app:survey}

This appendix gives the reading of 18 recent RLVR and reasoning-model releases behind the count in
Section~\ref{sec:disagree} and in Table~\ref{tab:disagree}, and sets the accuracies published for pairs we evaluate
beside our readings. For each release, the paper and, where one exists, its released evaluation code were read in
August 2026 for how each reported out-of-domain multiple-choice number was produced; the releases in which we first
found none were read again, in every version, in September 2026. A release is one paper or repository, so
SimpleRL-Zoo and SimpleRL-Zero, which share a repository, count as one. A result is a set of such numbers that one
release produces the same way; four releases produce different benchmarks, or different model stages, differently
and so contribute two results each. A result counts as generated when the paper or the code shows that the model
writes text from which the answer is read (a stated sampling temperature, or an average over sampled runs, counts
as showing this); as log-likelihood when the options are scored by their likelihood; and as unstated when neither
says.

Thirteen releases report at least one such number, in 17 results (Table~\ref{tab:app-survey}): 13 are generated, 1
is scored by log-likelihood (Tulu~3's headline MMLU, while the same paper's GPQA and MMLU-Pro numbers are
generated), and 3 are unstated. Five releases report none. Where these releases report an out-of-domain
multiple-choice number at all, it was almost always produced by generating text and reading the answer out of it,
so a format gain can arise in them as it does on our outputs. The count says nothing about how large the format
gain is in these releases, and the 18 are not a systematic sample: they were not chosen by a stated rule.

\begin{table}[!htbp]
\caption{\textbf{How 18 releases produced their out-of-domain multiple-choice numbers.} Benchmarks in
parentheses.}
\label{tab:app-survey}
\centering
\small
\begin{tabular}{p{0.14\textwidth}rp{0.70\textwidth}}
\toprule
produced by & results & releases \\
\midrule
generation & 13 & DeepSeek-R1 (MMLU, MMLU-Redux, MMLU-Pro, GPQA-Diamond, C-Eval) and its distilled models
(GPQA-Diamond); Qwen2.5-Math (MMLU-STEM); SimpleRL-Zoo (MMLU, MMLU-STEM, GPQA-Diamond); MiMo-7B-RL (GPQA-Diamond, MMLU-Pro); Open-Reasoner-Zero
(GPQA-Diamond); Light-R1 (GPQA-Diamond); OpenThinker3-7B (GPQA-Diamond, MMLU-Pro); s1 (GPQA-Diamond); Tulu~3
(GPQA, MMLU-Pro); Absolute Zero Reasoner (MMLU-Pro); Llama-Nemotron (MMLU, GPQA-Diamond); Phi-4-reasoning (MMLU,
MMLU-Pro, GPQA-Diamond) \\
log-likelihood & 1 & Tulu~3 (MMLU) \\
unstated & 3 & Kimi k1.5 (MMLU, C-Eval); MiMo-7B-Base (MMLU, MMLU-Pro, ARC-Challenge, GPQA-Diamond, BBH,
AGIEval, CMMLU, C-Eval); Open-Reasoner-Zero (MMLU, MMLU-Pro) \\
none reported & --- & QwQ-32B; DAPO; Skywork-OR1; AceReason-Nemotron; Eurus-2-7B-PRIME \\
\bottomrule
\end{tabular}
\end{table}

\paragraph{What is published for the pairs we evaluate.} The SimpleRL-Zoo paper reports MMLU, MMLU-STEM and
GPQA-Diamond accuracy for its base models and their RL models, and reads the gains as generalization
\citep[Table~2]{simplerlzoo}. It describes its evaluation as a script based on Qwen2.5-Math's, a sampling
temperature of 1.0, up to 16K tokens and the prompt template of training. Its repository ships the MMLU-STEM items
for that script and neither items nor a script for MMLU or GPQA-Diamond, and the paper does not say which decoding
this table used. We class the result as generated. The SimpleRL-Zero checkpoints have no report of their own.
General-Reasoner \citep{generalreasoner} reports MMLU-Pro for one of our pairs and does not say how answers were
read.

Table~\ref{tab:app-published} sets these accuracies beside three readings of our own outputs for the same pair and
benchmark. The two sides differ in items, prompt, decoding and extractor, so the table does not split any published
gain. Published out-of-domain multiple-choice gains exist for these checkpoints, including the pairs for which our
outputs show no gain to split. No single reading of our outputs reproduces SimpleRL-Zoo's accuracies: for
Qwen2.5-Math-7B, the release prints 38.0\% and 56.6\% on MMLU, where under the boxed-instruction
prompt our outputs give 53.6\% and 57.4\% by the fixed extractor and 24.0\% and 55.8\% by the boxed-only one.
The one number from an independent evaluation, General-Reasoner's gain of 3.8 points on MMLU-Pro, is of the size
that the fixed extractor reads on our outputs and far below the boxed-only reading. Splitting the published gains
would take outputs generated under the release's own prompt and decoding.

\begin{table}[!htbp]
\caption{\textbf{Published accuracies for pairs of this study, beside our readings of the same pair and
benchmark.} Published: the base model, its SimpleRL-Zoo model and the gain in points. Ours: the gain on our 500
items under the boxed-instruction and the plain prompt, as read by the fixed extractor, by the boxed-only MATH
extractor, and by Qwen2.5-Math's extractor, whose MMLU branch is the letter rule of the script that SimpleRL-Zoo's
evaluation is based on. $^{\dagger}$The fixed extractor's extraction rates do not meet the criterion of
Section~\ref{sec:criteria}.}
\label{tab:app-published}
\centering
\small
\setlength{\tabcolsep}{4pt}
\begin{tabular}{lrrrccc}
\toprule
 & \multicolumn{3}{c}{published} & \multicolumn{3}{c}{our outputs: gain, boxed-instruction / plain} \\
\cmidrule(lr){2-4}\cmidrule(lr){5-7}
base model & base & RL & gain & fixed extractor & boxed only & Qwen2.5-Math \\
\midrule
\multicolumn{7}{l}{\emph{MMLU; published in SimpleRL-Zoo, Table 2}} \\
Qwen2.5-Math-7B & 38.0 & 56.6 & $+18.6$ & $+3.8$ / $+3.0$ & $+31.8$ / $+22.0$ & $-12.2$ / $-12.0$ \\
Qwen2.5-7B & 38.6 & 47.0 & $+8.4$ & $+5.8$ / $+3.0$ & $+28.4$ / $+44.8$ & $-0.8$ / $+2.2$ \\
Mistral-7B-v0.1 & 28.0 & 34.6 & $+6.6$ & $+1.0$ / $-12.2$ & $-2.8$ / $0.0$ & $-4.6$ / $-8.2$ \\
DeepSeek-Math-7B & 22.7 & 45.5 & $+22.8$ & $+4.2^{\dagger}$ / $-2.0$ & $+2.6$ / $0.0$ & $-3.6$ / $+0.2$ \\
Qwen2.5-1.5B & 35.4 & 45.2 & $+9.8$ & $+0.8^{\dagger}$ / $-0.2^{\dagger}$ & $+1.6$ / $0.0$ & $+3.4$ / $+0.8$ \\
Mistral-Small-24B & 31.7 & 78.8 & $+47.1$ & $+25.0^{\dagger}$ / $+7.4$ & $+40.4$ / $0.0$ & $+4.6$ / $+0.8$ \\
\midrule
\multicolumn{7}{l}{\emph{MMLU-Pro; published in General-Reasoner, Table 2}} \\
Qwen2.5-7B & 47.7 & 51.5 & $+3.8$ & $+2.4$ / $-1.6$ & $+15.6$ / $+25.2$ & $+1.0$ / $+7.6$ \\
\bottomrule
\end{tabular}
\end{table}

\section{Cost and reproducibility}
\label{app:cost}

This appendix gives the cost of generation, the settings that fix every output and interval, the prompts, and
what the release holds.

\paragraph{Cost.} The 192 generation runs of the grid took 29.7 GPU-hours of generation and 1.5 hours of model
loading. Four of the 12 checkpoints took 89\% of the generation time: Mistral-7B, DeepSeek-Math-7B and their RL
models, which ship no chat template and run to the 2048-token limit on 61\% to 86\% of items. Scoring needs no
GPU: every step after generation runs on a CPU.

\paragraph{Settings.} Decoding is greedy, so no sampling seed enters generation. All models run in
\texttt{float16}, because not every GPU used supports \texttt{bfloat16}; both models of every pair use the same
precision. Each benchmark contributes 500
items, except AQuA-RAT and SAT-Math, whose 254 and 220 test items are used in full. Items were frozen and hashed
before any generation, and every generation file records the hash of the item file it was built from; all
records match. Every interval is seeded. Each comparison's bootstrap of 10,000 draws is seeded from a fixed seed
and a hash of the comparison's name, so that each interval is reproducible on its own; the pooled and paired
bootstraps over each benchmark's items use seed 20260816, the permutation test of Appendix~\ref{app:tracks} seed 20260818, and the readers'
sample and bootstrap seed 20260930. The interval of the pilot's format share in Appendix~\ref{app:omega} is an
exception: its seed includes a hash that Python salts per process, so it is the interval of the one recorded run,
and a new run reproduces the share and moves the interval's ends slightly.

\paragraph{Prompts.} Each prompt is wrapped in the checkpoint's own chat template where the checkpoint ships one.
The boxed-instruction prompt for multiple choice reads as follows, with its first line broken for print:
{\footnotesize
\begin{verbatim}
Answer the following multiple-choice question. Reason step by step,
then give your final answer as a single letter inside \boxed{}.

Question: <question>
A) <first option>
B) <second option>
...
\end{verbatim}
}
The plain prompt for multiple choice omits the second sentence. For free-form items, both prompts begin
``Solve the following problem.'' in place of the first sentence, write ``Problem:'' in place of ``Question:'',
list no options, and, in the boxed-instruction prompt, ask for the final answer ``inside \verb|\boxed{}|''
without ``as a single letter''.

\paragraph{Release.} The ancillary files of this arXiv version hold the analysis code, the 8 frozen item files with
their hashes, the result files, the records behind the counts of Appendices~\ref{app:population}, \ref{app:omega},
\ref{app:extractors} and~\ref{app:survey}, the independent readers' material, the written declarations of the readers'
check and of how the two larger pairs would be read, the Llama pilot's 8 generation files (two are second copies of its ARC-Challenge
files) with its item files, and the
SHA-256 hash of each of the 264 generation files (192 from the grid, 64 from the two larger pairs and 8 from the Llama
pilot). The other 256 generation files, which together exceed arXiv's size limit, are available from the author; each
can be checked against its hash. With them, any extractor, including one written after this paper, can be scored
against the same outputs without a GPU. The result files behind the paper's numbers that the letter step can change are held twice: in
\nolinkurl{results/with-letter-step/} as this paper reports them, written by \nolinkurl{harness/with_letter_step.py},
and in \nolinkurl{results/} with Qwen2.5-Math's extraction function alone. The others are held once, in
\nolinkurl{results/}.

\end{document}